\PassOptionsToPackage{numbers}{natbib}
\documentclass{article}
\usepackage[final]{neurips_2023}

\usepackage[utf8]{inputenc} %
\usepackage[T1]{fontenc}    %
\usepackage{hyperref}       %
\usepackage{url}            %
\usepackage{booktabs}       %
\usepackage{amsfonts}       %
\usepackage{nicefrac}       %
\usepackage{microtype}      %
\usepackage[dvipsnames]{xcolor}

\usepackage{wrapfig,lipsum,booktabs}
\usepackage{graphicx} %
\usepackage{amsmath}
\usepackage{amsthm}
\usepackage{amsfonts}
\usepackage{caption}
\usepackage{subcaption}
\usepackage{xcolor}
\usepackage{amsmath}
\usepackage{amssymb, bm}
\usepackage{amsbsy}
\usepackage{arydshln}
\usepackage{mathtools}

\usepackage{hyperref}
\hypersetup{
	colorlinks=true,
	linkcolor=blue,
	filecolor=magenta,
	urlcolor=magenta,
}

\newtheorem{theorem}{Theorem}[section]

\newtheorem{lemma}[theorem]{Lemma}

\newtheorem{definition}[theorem]{Definition}

\newtheorem{remark}[theorem]{Remark}

\usepackage{algorithm}
\usepackage{algpseudocode}
\usepackage{multirow}
\usepackage{booktabs}

\title{Primal-Attention: Self-attention through  
	Asymmetric Kernel SVD in Primal Representation}

\author{
	Yingyi Chen
	\thanks{Equal contribution.}\,
	\\
	ESAT-STADIUS \\
	KU Leuven, Belgium \\
	\texttt{yingyi.chen@esat.kuleuven.be} \\
	\And
	Qinghua Tao 
	\footnotemark[1]\,
	\\
	ESAT-STADIUS \\
	KU Leuven, Belgium \\
	\texttt{qinghua.tao@esat.kuleuven.be} \\
	\And
	Francesco Tonin \\
	ESAT-STADIUS \\
	KU Leuven, Belgium \\
	\texttt{francesco.tonin@esat.kuleuven.be} \\
	\And
	Johan A.K.~Suykens \\
	ESAT-STADIUS \\
	KU Leuven, Belgium \\
	\texttt{johan.suykens@esat.kuleuven.be} \\
}

\begin{document}

	\maketitle

	\begin{abstract}
		Recently, a new line of works has emerged to understand and improve self-attention in Transformers by treating it as a kernel machine. 
		However, existing works apply the methods for symmetric kernels to the asymmetric self-attention, resulting in a nontrivial gap between the analytical understanding and numerical implementation. 
		In this paper, we provide a new perspective to represent and optimize self-attention through asymmetric Kernel Singular Value Decomposition (KSVD), which is also motivated by the low-rank property of self-attention normally observed in deep layers.
		Through asymmetric KSVD, 
		\emph{i)} a primal-dual representation of self-attention is formulated, where the optimization objective is cast to maximize the projection variances in the attention outputs;
		\emph{ii)} a novel attention mechanism, i.e., Primal-Attention, is proposed via the primal representation of KSVD, avoiding explicit computation of the kernel matrix in the dual; 
		\emph{iii)}
		with KKT conditions, we prove that the stationary solution to the KSVD optimization in Primal-Attention yields a {zero-value} objective. In this manner, KSVD optimization can be implemented by simply minimizing a regularization loss, so that low-rank property is promoted without extra decomposition.
		Numerical experiments show state-of-the-art performance of our Primal-Attention with improved efficiency. 
		Moreover, we demonstrate that the deployed KSVD optimization regularizes Primal-Attention with a sharper singular value decay than that of the canonical self-attention, further verifying the great potential of our method.
		To the best of our knowledge, this is the first work that provides a \emph{primal-dual representation} for the \emph{asymmetric kernel} in self-attention and successfully applies it to \emph{modelling} and \emph{optimization}\footnote{Our implementation is available at~\url{https://github.com/yingyichen-cyy/PrimalAttention}}.

	\end{abstract}

	\section{Introduction}
	\label{sec::intro}

	Transformers~\cite{vaswani2017attention} have become ubiquitous nowadays with state-of-the-art results in various tasks, such as natural language processing~\cite{kenton2019bert,brown2020language,raffel2020exploring}, computer vision~\cite{fan2021multiscale,liu2021swin,touvron2021training,chen2023jigsaw}, reinforcement learning~\cite{janner2021offline,chen2021decision,wu2022flowformer}, etc. 
	In the remarkable success of Transformers, the self-attention blocks play a key role, where the complicated dependencies between {the individuals} in data sequences can be depicted by using the established queries, keys, and values. 
	Despite the prevailing advantages, 
	theoretical understandings towards Transformers seem yet lagged behind its unprecedented empirical performance. 
	
	Recently, the kernel-based perspective has been proposed 
	{where the dot-product attention operation is shown} as a kernel matrix~\cite{tsai2019}. 
	This finding is quite encouraging by bridging kernels~\cite{vapnik1999overview} with Transformers, as kernel methods have long been well studied with good interpretation ability. 
	Following this spirit, different works have been proposed subsequently to improve self-attention, e.g.,~\cite{choromanski2021rethinking,ali2021xcit,nguyen2022improving,nguyen2022fourierformer,chi2022kerple,nguyen2023a}. 
	However, in these works, the applied kernel techniques rely on Mercer kernels~\cite{mercer1909}  requesting symmetry, which is inconsistent with the 
	{intrinsically asymmetric} setups of self-attention. 
	In~\cite{wright2021transformers}, it analytically characterizes the attention by asymmetric kernels based on Reproducing Kernel Banach Spaces (RKBS)~\cite{zhang2009reproducing}.
	{Nonetheless,} neither the asymmetry property nor the related optimization is utilized for improvements. 
	In~\cite{nguyen2023a}, self-attention is derived with a primal-dual representation from the support vector regression, which still adopts the technique for Mercer kernels.
	Moreover, in the cast supervised task, the assumed ground-truth  outputs of self-attention are practically non-existent, making it difficult to be applied in the optimization.
	
	\begin{figure}[t]
		\centering    
		{\includegraphics[width=\textwidth]{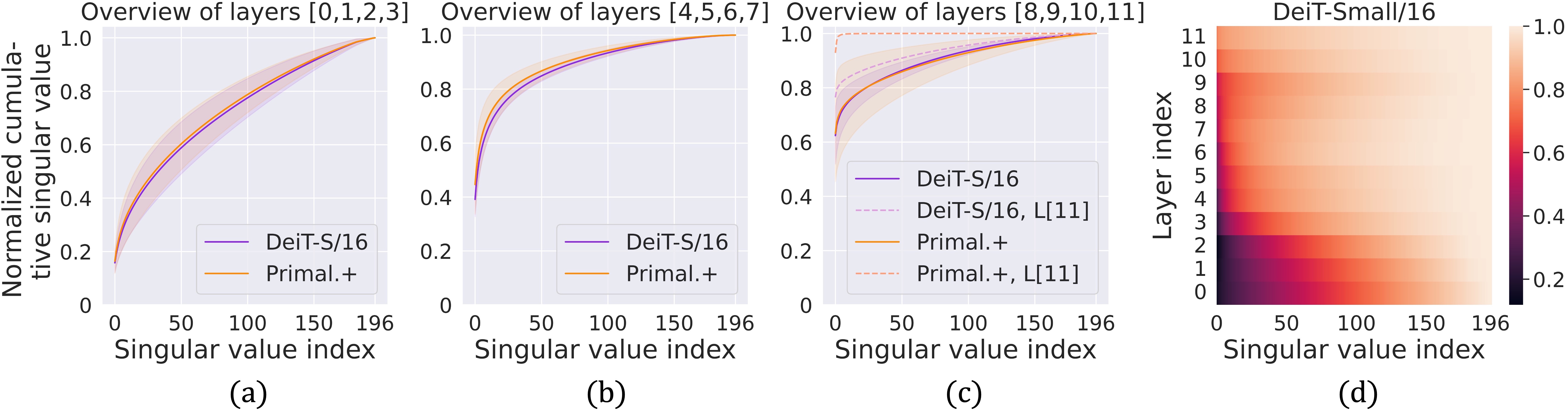}}
		\caption{
			Spectrum analysis
			of the self-attention matrix on ImageNet-1K~\cite{deng2009imagenet}.
			(a)-(c) plot the cumulative explained variance regarding the singular values of the attention matrix with mean and standard deviation of the chosen layers in pre-trained DeiT-Small/16~\cite{touvron2021training} and Primal.$+$DeiT-Small/16 (ours): the attention matrix attains sharper singular value decays in deeper layers, also shown in (d).
			{Note that we also plot the cumulative explained variance curves of the self-attention matrix from the last layer, i.e.,~the 11-th layer denoted by ``L[11]'', of both models in (c).}
			Our method shows an enhanced low-rank property of the attention matrix upon the baseline.
		}
		\label{fig:overview:low-rank}
	\end{figure}

	In this work, we provide a novel perspective to interpret self-attention with a primal-dual representation based on asymmetric Kernel Singular Value Decomposition (KSVD), which fills
	the gap of dismissing the asymmetry between theory and implementation. 
	{Specifically, in this unsupervised setup, we propose to remodel self-attention in the primal representation, namely, Primal-Attention, and to optimize it accordingly.}
	{Our method is driven by two major motivations.}
	{Firstly, 
		we observe that attention matrices in Transformers can be
		low-rank, as shown in Figure~\ref{fig:overview:low-rank}(d), and this property becomes more significant towards deeper network layers.}
	{Secondly, the self-attention matrix is intrinsically an asymmetric kernel matrix~\cite{tsai2019,wright2021transformers}.}
	To this end, we propose KSVD for self-attention, which takes both low-rank and asymmetric properties into consideration. 
	To the best of our knowledge, this is the first work that provides a primal-dual representation for the asymmetric self-attention and applies it to modelling and optimization.
	The  contributions of this work are summarized as follows:
	\begin{itemize}
		\item We characterize self-attention by KSVD with asymmetric kernels. Different from existing works employing symmetric kernel-based methods, we take asymmetry into account so as to be more consistent with the real setups in self-attention. (Section~\ref{sec::problem_statesment})
		
		\item  We derive a primal-dual representation for self-attention {through KSVD}, and {propose} a novel attention in the primal, named Primal-Attention, avoiding the expensive kernel computation in the dual.
		With KSVD, the values are interpreted as the projection weights that yield maximal variances of features, and  the low-rank property can be 
		pursued by confining the projection numbers. 
		(Section~\ref{sec::primal_dual}, Section~\ref{sec::method})
		
		\item  We prove that the stationary solution to the derived KSVD leads to a zero-value objective 
		of the unconstrained primal problem.
		Therefore, the optimization of KSVD in Primal-Attention can be efficiently implemented by minimizing a regularization term added to the loss, with no need of extra decomposition operations. 
		(Section~\ref{sec::method})

		\item In numerical experiments, Primal-Attention achieves state-of-the-art performance on various datasets together with efficiency advantages over the canonical self-attention. 
		Moreover, we demonstrate that our deployed optimization from KSVD can regularize the attention with a sharper singular value decay, hence promoting learning more low-rank features, which is shown in Figure~\ref{fig:overview:low-rank}. (Section~\ref{sec::exp})

	\end{itemize}

	\section{Problem Statement: Self-attention with Asymmetric Kernel}
	\label{sec::problem_statesment}

	\paragraph{Self-attention}
	Let $\{\bm{x}_i \in\mathbb{R}^{d}\}_{i=1}^N$ be the input data sequence. 
	In self-attention~\cite{vaswani2017attention},
	the queries, keys and values output the linear projections of the input sequence, such that
	\begin{align}\label{eq:q:k:v}
		q(\bm {x}_i) = W_q \bm x_i,
		\quad
		k(\bm {x}_i) = W_k \bm x_i,
		\quad
		v(\bm {x}_i) = W_v \bm x_i,
	\end{align}
	where $W_q\in\mathbb{R}^{d_q\times d}$, $W_k\in\mathbb{R}^{d_k\times d}$, and $W_v\in\mathbb{R}^{d_v\times d}$, commonly with the setup $d_q=d_k$.
	The attention scores are then given by 
	$a (\bm x_i, \bm x_j) ={\left < q(\bm x_i),  k(\bm x_j) \right>}/ {\sqrt{d_k}} = \left <  W_q \bm x_i,  W_k \bm x_j \right> / {\sqrt{d_k}}$.
	In the canonical self-attention, the ``softmax'' activation is then applied to bring non-linearity and positives, yielding the attention weights:
	\begin{align} \label{eq::attnweight}
		\kappa (\bm x_i, \bm x_j)
		& =\text{softmax}\left( \left <  W_q \bm x_i,  W_k \bm x_j \right > / {\sqrt{d_k}}\right),
		\quad i,j = 1,\ldots,N.
	\end{align}
	Similar to~\cite{tsai2019}, the attention matrix, i.e., $K \coloneqq  [\kappa(\bm x_i, \bm x_j)] \in \mathbb R^{N\times N}$, can be interpreted as a kernel matrix with entries $\kappa (\bm x_i, \bm x_j)$, where $\kappa (\cdot, \cdot)\colon \mathbb R^{d} \times \mathbb R^d \to \mathbb R$ serves as the kernel function. 
	Notice that in general, $\left <  W_q \bm x_i,  W_k \bm x_j \right> \neq \left<  W_q \bm x_j,  W_k \bm x_i \right>$, leading to an asymmetric kernel where $K_{ij} \neq K_{ji}$.

	Then, the attention output $\bm o_i \in \mathbb R^{d_v}$ in each head is attained as:
	\begin{align} \label{eq::dual_output}
		\bm  o_i = \sum\nolimits_{j=1}^N v(\bm x_j)\kappa (\bm x_i, \bm x_j) = \sum\nolimits_{j=1}^N v(\bm x_j)K_{ij},
		\quad i=1,\ldots,N.
	\end{align}
	In Transformers, multiple heads are commonly applied through the concatenation of all heads~\cite{vaswani2017attention}.

	\paragraph{Asymmetric Attention Matrix} 
	In kernel methods, rigorous works have been presented with Mercer kernels that are symmetric and positive semi-definite~\cite{mercer1909} through the kernel trick from Reproducing Kernel Hilbert Spaces (RKHS)~\cite{vapnik1999overview}. 
	On the other hand in Transformer~\cite{vaswani2017attention}, the attention kernel matrix is asymmetric as shown in \eqref{eq::attnweight}.  
	Existing works leverage the kernel interpretation for improving self-attention~\cite{choromanski2021rethinking,nguyen2022fourierformer,chi2022kerple,nguyen2023a}, however, their deployed kernel-based techniques all rely on Mercer kernels, which is inconsistent with the asymmetric {nature}.  
	Instead, asymmetry is allowed in kernel tricks from Reproducing Kernel Banach Spaces (RKBS)~\cite{zhang2009reproducing} as in the following Definition~\ref{def:asym:kernel}.

	\begin{definition}
		[Definition 2~\cite{wright2021transformers}; Theorem 2.1~\cite{lin2022reproducing};\cite{georgiev2013construction}]
		\label{def:asym:kernel}
		For asymmetric kernels, the kernel trick from RKBS with the kernel function $\kappa(\cdot, \cdot)\colon \mathcal X \times \mathcal Z  \to \mathbb R$ can be defined by the inner product of two real measurable feature maps from Banach spaces $ \mathcal B_{\mathcal X},  \mathcal B_{\mathcal Z}$ on $\mathcal{X}, \mathcal{Z}$, respectively:
		\begin{equation}\label{eq:def:kernel}
			\kappa( \bm x, \bm z) = \left< \phi_x( \bm x), \phi_z( \bm z) \right>,\, \forall  \bm x \in \mathcal X,\, \phi_x \in \mathcal B_{\mathcal X},  \bm z \in \mathcal Z, \phi_z \in \mathcal B_{\mathcal Z}.
		\end{equation}
	\end{definition}
	Based on Definition~\ref{def:asym:kernel}, the kernel matrix in self-attention can be characterized by the kernel trick from RKBS~\cite{wright2021transformers}, providing an analytical tool from the aspect of kernel representer theorem.
	
	\paragraph{SVD and Shifted Eigenvalue Problem}
	SVD factorizes a given $r$-rank matrix $A\in \mathbb R^{N\times M}$ by two sets of orthonormal eigenbases:  $  A = U \Sigma   V^\top $ with  $ \Sigma =\text{diag} \{\sigma_1, \ldots, \sigma_r\}$ of positive singular values and the columns of $ U \in \mathbb{R}^{N \times r}$ and $ V \in \mathbb{R}^{M \times r}$ as the \emph{left} and \emph{right singular vectors}, respectively~\cite{strang2006linear}. 
	$U$, $V$ reflect the subspace projections in relation to the columns and rows, as shown in  \eqref{shifted:eigen}, and contain different information residing in $A$ due to the asymmetry. 
	When $A$ is squared and symmetric, SVD boils down to the eigendecomposition with $U= V$.  
	In~\cite{suykens2016svd}, a novel variational principle is proposed for SVD with Least Squares Support Vector Machines (LSSVM)~\cite{suykens2002least}, where the dual problem leads to a shifted eigenvalue problem in accordance to the decomposition theorem from Lanczos~\cite{lanczos1958linear} regarding SVD, i.e., Theorem \ref{theorem:lanczos}. This theorem is also of special importance in our work to the kernel extension of SVD  in self-attention under the framework of LSSVM.
	
	\begin{theorem}[Lanczos~\cite{lanczos1958linear}]\label{theorem:lanczos}
		Any non-zero matrix {$A \in \mathbb{R}^{N \times M}$} can be written as {$ A=\tilde{ U}\tilde{ \Sigma}\tilde{ V}^\top$}, where the matrices $\tilde{ U}$, $\tilde{ \Sigma}$, $\tilde{V}$ are defined by the shifted eigenvalue problem:
		\begin{equation}\label{shifted:eigen}
			\begin{aligned}
				A \tilde{ V}& =  \tilde{ U}\tilde{ \Sigma}, \\
				{ A}^\top\tilde{ U}& =  \tilde{ V}\tilde{ \Sigma}, 
			\end{aligned}
		\end{equation}
		where  $\tilde{ U}\in \mathbb R^{N\times r}$ and  $\tilde{ V}\in \mathbb R^{M\times r}$ satisfy $\tilde{ U}^\top\tilde{ U}= I_r$ and $\tilde{ V}^\top\tilde{ V}= I_r$, and $\tilde{ \Sigma}\in \mathbb R^{r\times r}$ is a diagonal matrix with positive numbers.
	\end{theorem}

	\section{Primal-dual Representation of Self-attention based on Kernel SVD}
	\label{sec::primal_dual}
	In this section, we apply the kernel trick from RKBS to the asymmetric attention kernel, and derive self-attention with a primal-dual representation based on 
	Kernel SVD (KSVD). 
	Under this learning scheme, a new self-attention mechanism is proposed by remodelling the attention output in the primal representation, without explicit computation of the kernel matrix in the dual representation. 
	With the stationarity conditions, we flexibly implement the optimization of KSVD through an additional loss term, which can regularize the model to improved low-rank properties without extra decomposition.
	
	\paragraph{{KSVD optimization problem} in Primal and Dual}
	By Definition \ref{def:asym:kernel} of RKBS, the kernel function in the dual for the asymmetric attention kernel $K$ in self-attention can be formulated by 
	$K_{ij} = \kappa (\bm x_i, \bm x _j) \coloneqq \left< \phi_q(\bm x_i), \phi_k(\bm x_j)\right>$,
	with two feature maps $\phi_q$, $\phi_k$ related to queries and keys. 
	Recall the self-attention output in \eqref{eq::dual_output}, the values {$\{v(\bm{x}_j)\}_{j=1}^N$ resemble the dual variables} projecting the kernel matrix in the dual representation of kernel methods, whereas the kernel involved is asymmetric.
	In this regard, the nonlinear version of SVD under the framework of LSSVM~\cite{suykens2016svd} well fits the self-attention setup, hence is set as the basis for the following work with asymmetric kernels built upon RKBS. 
	{Differently, we extend the matrix SVD setups in~\cite{suykens2016svd} to the case of asymmetric attention matrix with two input data sources as queries and keys.} 
	Moreover,  we consider that in self-attention, values are input data-dependent, and thus generalize~\cite{suykens2016svd} with data-dependent projection weights.
	
	Given the sequence  $\{\bm x_i \in \mathbb{R}^d\}_{i=1}^N$, we start from the  primal optimization with KSVD:
	\begin{equation}\label{eq:ksvd:lssvm:std:attention}
		\begin{array}{rl}
			\max\limits_{W_e, W_r, \bm e_i, \bm r_j }&  J =  \dfrac{1}{2}\sum\nolimits_{i=1}^N \bm e_i^\top\Lambda\bm e_i + \dfrac{1} {2}\sum\nolimits_{j=1}^N  \bm r_j^\top \Lambda \bm r_j - \text{Tr}\left(W_e^\top W_r\right)  \\
			{\rm s.t.} &  
			{\bm e_i = (f(X)^\top W_e)^\top \phi_q(\bm x_i), \  i=1, \ldots, N,}
			\\
			& 
			{\bm r_j = (f(X)^\top W_r)^\top \phi_k(\bm x_j), \ j=1, \ldots, N,}
		\end{array}
	\end{equation}
	where we have the data-dependent projection weights $f(X)^\top W_e =: W_{e|X} \in \mathbb{R}^{p\times s}$, $f(X)^\top W_r =: W_{r|X} \in \mathbb{R}^{p\times s}$ relying on parameters $W_e, W_r \in \mathbb{R}^{N \times s}$, 
	the feature maps
	$\phi_q(\cdot), \phi_k(\cdot)\colon \mathbb{R}^d \to \mathbb{R}^p$, 
	the projection scores $\bm{e}_i, \bm{r}_j \in \mathbb{R}^{s}$,
	and the regularization coefficient $\Lambda \in \mathbb{R}^{s\times s}$ which is a positive diagonal matrix. %
	The objective $J$ in the primal optimization 
	maximizes the projection variances of $W_{e|X}^\top\phi_q(\bm{x}_i)$, $W_{r|X}^\top\phi_k(\bm{x}_j)$ regarding queries and keys, and also involves a regularization term coupling the projections.
	The corresponding solution in the dual is characterized
	by the right and left singular vectors, which captures the directions with maximal projection variances, w.r.t.~rows (queries) and columns (keys) of the attention kernel matrix.
	Thus, with the formulated primal optimization in \eqref{eq:ksvd:lssvm:std:attention},  the learning in self-attention is interpreted  as a SVD problem on the attention matrix.
	
	For clarity, we elaborate our primal optimization problem in \eqref{eq:ksvd:lssvm:std:attention} as follows.
	\textit{i)} The projection weights are data-dependent, where 
	{$f(X)=:F_X \in \mathbb{R}^{N \times p}$} denotes a transformation matrix containing the information of the sequence data $X\coloneqq [\bm{x}_1,\ldots,\bm{x}_N]^\top\in \mathbb{R}^{N\times d}$.
	Notably, ${F_X}$ is a constant matrix once given $X$, and we let it linearly depend on $X$ in experiments.
	Moreover, when ${F_X}$ is chosen as the identity matrix, it reconciles to the common setups in kernel methods in primal.
	\textit{ii)} The feature maps related to queries and keys, respectively, are  defined as $\phi_q(\bm{x}_i)\coloneqq g_q(q(\bm{x}_i))$, $\phi_k(\bm{x}_i)\coloneqq g_k(k(\bm{x}_i))$, where $g_q(\cdot):\mathbb{R}^{d_q}\to \mathbb{R}^{p}$ and $g_k(\cdot):\mathbb{R}^{d_k}\to \mathbb{R}^{p}$ denote the mappings composited on the linear projections $q(\cdot)$ and $k(\cdot)$ in \eqref{eq:q:k:v} of queries and keys.
	Note that we leave the choice of $g_q$ and $g_k$ later explained in Remark~\ref{rkm::feat_maps}.
	\textit{iii)} By projecting $\phi_q(\bm{x}_i), \phi_k(\bm{x}_j) \in \mathbb{R}^p$ with  weights $W_{e|X}, W_{r|X} \in \mathbb{R}^{p\times s}$, we obtain the projection scores $\bm{e}_i, \bm{r}_j \in \mathbb{R}^s$ along the $s$ directions, {usually $s < p$}, which corresponds to the number of singular values of the induced kernel matrix in the dual optimization \eqref{shifted:eigen:ksvd:std:attention}.

	\begin{remark} [Variance maximization objective]\label{rmk:obj}
		In the formulated KSVD problem, the objective in the primal optimization \eqref{eq:ksvd:lssvm:std:attention} jointly maximizes the variances of the two 
		{projections $\bm{e}_i$, $\bm{r}_j$ in the feature spaces determined by $\phi_q$, $\phi_k$ along the directions $W_{e|X}$, $W_{r|X}$.}
	Within this context,  $\bm e_i$, $\bm r_j$ 
	learn to capture maximal information mutually residing in $\phi_q$ and $\phi_k$ regarding the queries and keys. 
\end{remark}

With Lagrangian duality and KKT conditions, we prove that the dual optimization problem to \eqref{eq:ksvd:lssvm:std:attention} leads to a shifted eigenvalue problem corresponding to the SVD on the asymmetric attention kernel $K$, given by Theorem \ref{theorem:ksvd:dual}. 
The proof is provided in the Supplementary Material.

\begin{theorem} [Dual optimization problem of KSVD in self-attention]\label{theorem:ksvd:dual}
	With Lagrangian duality and the KKT conditions, the dual optimization problem of \eqref{eq:ksvd:lssvm:std:attention} leads to the shifted eigenvalue problem:
	\begin{equation}\label{shifted:eigen:ksvd:std:attention}
		\begin{aligned}
			K H_r & = H_e {\Sigma}, 
			\\
			K^\top H_e &= H_r {\Sigma},
		\end{aligned}
	\end{equation}
	where $\Sigma \in \mathbb R^{s\times s}$ is a positive diagonal matrix, and  $H_e = [\bm h_{e_1}, \ldots, \bm h_{e_N}]^\top \in \mathbb R^{N\times s}$, 
	$H_r = [\bm h_{r_1}, \ldots, \bm h_{r_{N}}]^\top \in \mathbb R^{{N}\times s}$ are the dual variables serving as the left and right singular vectors, respectively.  
	{The kernel trick to the asymmetric kernel matrix $K$ is interpreted as $K_{ij}=\left< f(X) g_q (q(\bm x_i)), f(X) g_k(k(\bm x_j))  \right> {=:} \left< \phi_q'(\bm x_i), \phi_k'(\bm x_j) \right>$.}
\end{theorem}

As shown in Theorem \ref{theorem:ksvd:dual}, the solutions collect non-zero $\Lambda$ in \eqref{eq:ksvd:lssvm:std:attention} such that $\Sigma  =\Lambda^{-1}$. 
Based on Lanczos' decomposition in Theorem \ref{theorem:lanczos}, we can then associate $\Sigma$ with the non-zero singular values of the attention kernel $K$ in \eqref{shifted:eigen:ksvd:std:attention}, and  $H_e, H_r$ with the left and right singular vectors of $K$, such that $K = H_e \Sigma H_r^\top$. 
Therefore, formulas \eqref{eq:ksvd:lssvm:std:attention} and \eqref{shifted:eigen:ksvd:std:attention} provide the optimization problems of performing KSVD with the attention kernel matrix $K$ in the primal and in the dual, respectively.

\paragraph{Self-attention as KSVD dual representation}
Firstly, we provide the primal-dual model representation of the derived KSVD  problem.
Secondly, we show that the dual representation of the model corresponds to the canonical self-attention.
The derivation details involving the KKT conditions are provided in the Supplementary Material.

\begin{remark}[Primal-dual representations of KSVD in self-attention] \label{rmk:primal:dual}
	In the KSVD formulations for the asymmetric kernel matrix in self-attention, with KKT conditions, the projection scores can be either represented in the primal using explicit feature maps or in the dual using  kernel functions:
	\begin{equation}\label{eq:primal:dual:ksvd:std:attention}
		\begin{array}{rll}
			\text{Primal:} & 
			\left\{
			\begin{array}{l}
				{e}(\bm{x}) = W_{e|X}^\top  \phi_q(\bm{x})
				\\
				{r}(\bm{x}) = W_{r|X}^\top   \phi_k(\bm{x})
			\end{array},
			\right.
			\quad
			\text{Dual:} &
			\left\{
			\begin{array}{l}
				{e}(\bm{x})  = \sum\nolimits_{j=1}^N   \bm  h_{r_j}    \kappa(\bm{x},\bm{x}_j) 
				\\
				{r}(\bm{x}) = \sum\nolimits_{i=1}^N  \bm   h_{e_i}    \kappa(\bm{x}_i,\bm{x}).
			\end{array}
			\right.
		\end{array}
	\end{equation}
\end{remark}

\begin{remark} [Correspondence of  KSVD and  canonical self-attention output] \label{rmk:e:score:self:attention}
	Recall the output $\bm{o}_i$ of canonical self-attention \eqref{eq::dual_output}, 
	it corresponds to the dual representation of the projection score $ e(\bm x)$ in  \eqref{eq:primal:dual:ksvd:std:attention}, i.e., $\bm o_i \triangleq e(\bm x_i)$.
	When the values {$\{v(\bm x_j)\}_{j=1}^N$} in canonical self-attention are chosen as the dual variables {$\{\bm h_{r_j}\}_{j=1}^N$}, i.e., {$\bm h_{r_j}\coloneqq v(\bm x_j)$, $j=1,\ldots,N$}, the values {$\{v(\bm x_j)\}_{j=1}^N$} play the role of the right singular vectors of $K$.
\end{remark}

From the perspective in Remark \ref{rmk:e:score:self:attention}, the  
{optimization goal} in self-attention is interpreted to jointly capture the maximal variances of $\bm e_i$, $\bm r_j$ as in \eqref{eq:ksvd:lssvm:std:attention}, {where the projection scores can be denoted as $\bm{e}_i:=e(\bm{x}_i)$, $\bm{r}_j:=r(\bm{x}_j)$ through the representations  \eqref{eq:primal:dual:ksvd:std:attention}.} 
{However,} the canonical self-attention only outputs the $\bm e_i$-score ($\bm o_i \triangleq \bm e_i$). 
In this sense, the output of the canonical self-attention only considers the {projection scores} involving the right singular vectors of the asymmetric attention kernel $K$.

\begin{figure}[t]
	\centering    
	{\includegraphics[width=0.98\textwidth]{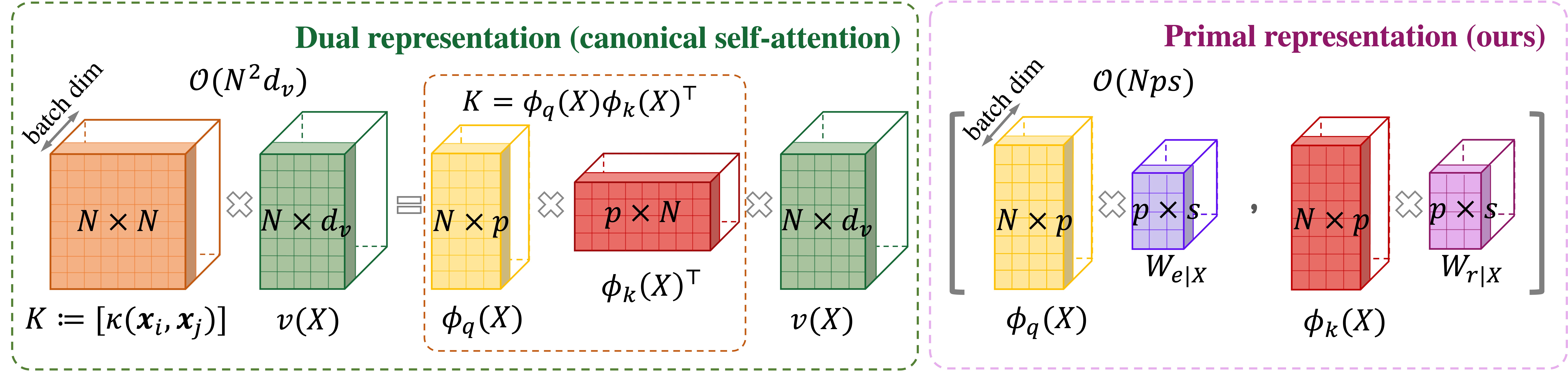}}
	\caption{An illustration of Primal-Attention and canonical self-attention.
		Left: in \emph{canonical self-attention}, the asymmetric attention matrix $K$ can be induced by two feature maps $\phi_q$, $\phi_k$ through kernel trick from RKBS. 
		The values $v(X)$ serve as the dual variables projecting the kernel matrix $K$ to the attention output. 
		The time and space complexity are {$\mathcal{O}(N^2d_v)$ and $\mathcal{O}(N^2 + Nd_v)$}. 
		Right: in our \emph{Primal-Attention}, we use the two feature maps $\phi_q$, $\phi_k$ in the primal to present the attention outputs, which are projected through the primal variables $W_{e|X}$, $W_{r|X}$. 
		The time and space complexity are $\mathcal{O}(Nps)$ and $\mathcal{O}(2Np + 2ps)$.
		To align with the output size in the canonical self-attention, we add a linear map mapping from $2s$ to $d_v$ with negligible memory increase after Primal-Attention's output.}
	\label{fig::pipeline}
	\vspace{-3mm}
\end{figure}

\section{Primal-Attention}
\label{sec::method}
\paragraph{Modelling}
It is quite remarkable that the attention output can be equivalently represented without the kernel expressions, avoiding the heavy computation of the kernel matrix. 
{Within} the context of KSVD, we further observe that there exists another set of projections $\bm r_j$ regarding the left singular vectors in $\bm h_{e_i}$ as  in \eqref{eq:primal:dual:ksvd:std:attention}, providing extra information residing in the asymmetric kernel matrix $K$.

We derive a novel attention mechanism by leveraging the primal representation of KSVD, namely, Primal-Attention, where two explicit feature maps $\phi_q, \phi_k$ are adopted. 
To fully exploit the asymmetry in the kernel matrix of self-attention, Primal-Attention concatenates the two sets of projections using both left and right singular vectors, and thus formulates the attention outputs {as follows:}
\begin{equation}\label{eq:outputs:Primal-Attention}
	\bm o_i 
	\coloneqq  [\bm e_i; \bm r_i]  
	= \left[  {W_{e|X}^\top}  \phi_q(\bm {x}_i); 
	{W_{r|X}^\top}  \phi_k(\bm{x}_i) \right] 
	= \left[  W_e^\top f(X) g_q (q(\bm x_i));  W_r^\top f(X) g_k (k(\bm x_i))\right].
\end{equation}
In Primal-Attention, the projection weights $W_{e|X}, W_{r|X}$ in the primal play the role as the counterparts of the values in the dual. 
{Given ${f(X)}=:F_X$ an identity matrix, our KSVD problem in \eqref{eq:ksvd:lssvm:std:attention} boils down to the data-independent projection weight case as in \cite{suykens2016svd}, which can thereby be regarded as a special case of our derived KSVD in Primal-Attention.
	In this case, the kernel trick in the asymmetric attention kernel becomes $K_{ij}=\left< g_q (q(\bm x_i)), g_k(k(\bm x_j))  \right> = \left< \phi_q(\bm x_i), \phi_k(\bm x_j) \right>$, $i,j = 1, \ldots, N$.}

\begin{remark}[Choices of  $\phi_q, \phi_k$ for non-linearity]\label{rkm::feat_maps}
	The canonical self-attention {adopts softmax for introducing non-linearity to the  attention score matrix}; {within our setups and kernel trick,} it can be viewed as: $\kappa(\bm{x}_i, \bm{x}_j)=\hat{D}^{-1}
	\left< \phi_q(\bm{x}_i), \phi_k(\bm{x}_i)\right>$ 
	where 
	$\phi_q(\bm{x}):=g(q(\bm{x}))$, $\phi_k(\bm{x}):=g(k(\bm{x}))$ with
	$g(\bm{z}):=\exp(-\|\bm{z}\|^2/2)(\exp(\bm{w}_1^\top \bm{z}),\ldots,\exp(\bm{w}_p^\top \bm{z}))$,
	$\bm{w}_i\sim \mathcal{N}(0, I_{d_q})$ and $d_q=d_k$
	~\cite{choromanski2021rethinking},
	and $\hat{D}:={\rm{diag}}( \phi_q(X) (\phi_k(X)^\top \bm{1}_N) )$.
	The projection scores then correspond to 
	$e(\bm{x})={\hat{D}^{-1/ 2}}W_{e|X}^\top \phi_q(\bm{x})$ and
	$r(\bm{x})={\hat{D}^{-1/ 2}}W_{r|X}^\top \phi_k(\bm{x})$.
	In this case, two exponential feature maps need to be constructed and the normalization factor $\hat{D}$ to all samples needs to be computed.
	In this paper, we consider feature maps related to the Cosine similarity kernel on queries and keys, such that
	$\phi_q(\bm{x}) = g_q(q(\bm x)) := q(\bm{x})/\|q(\bm{x})\|_2$ and 
	$\phi_k(\bm{x}) = g_k(k(\bm x)) := k(\bm{x})/\|k(\bm{x})\|_2$ in all experiments. 
	It is easy to implement and able to bring both non-linearity and normalization to the feature maps.
\end{remark}

\paragraph{Optimization}
The KSVD problem can either be optimized in the primal as a constrained optimization problem, or in the dual as a shifted eigenvalue problem (SVD on the kernel matrix $K$).  
In Primal-Attention, we perform the optimization in the primal.
In the following Lemma \ref{prop:obj:0}, we prove the {zero-value} property of the primal optimization objective $J$ in  \eqref{eq:ksvd:lssvm:std:attention} when it is evaluated at the stationary solution in \eqref{shifted:eigen:ksvd:std:attention}. 
The proof is provided in the Supplementary Material.

\begin{lemma}[A {zero-value} objective with stationary solutions]\label{prop:obj:0}
	The solutions of $H_e, H_r, \Sigma$ to the shifted eigenvalue problem in the dual optimization \eqref{shifted:eigen:ksvd:std:attention} lead to the  {zero-value} objective $J$ in the primal optimization \eqref{eq:ksvd:lssvm:std:attention}.
\end{lemma}

With the property in Lemma \ref{prop:obj:0}, rather than solving SVD problem on the kernel matrix $K$ in the dual, the optimization of Primal-Attention is realized by minimizing the primal objective to zero:
\begin{align} \label{eq::objective_function}
	\min  \ L + \eta 
	{\sum\nolimits_{l}  J_{l}^2},
\end{align}
where $L$ is the task-oriented loss, e.g., the cross-entropy loss for classification tasks, the summation term over $l$ denotes the additive objectives $J_l$ of all attention blocks using the proposed Primal-Attention, {where $J_l$ is implemented as the mean over all heads,} {and {$\eta >0$} is the regularization coefficient.}
Specifically, for each head in the self-attention using Primal-Attention, we have
\begin{equation}\label{eq:obj:std}
	\begin{array}{rl}
		& J({W_e, W_r}, \Lambda) =  \frac{1} {2}\sum\nolimits_{i=1}^N \bm e_i^\top\Lambda\bm e_i + \frac {1}  {2}\sum\nolimits_{j=1}^N  \bm r_j^\top \Lambda \bm r_j - \text{Tr}\left(W_e^\top W_r\right)  \\
		& =    \frac{1}{2}\sum\nolimits_{i=1}^N \|( W_{{e|X}}\Lambda^{\frac{1}{2}})^\top \phi_q(\bm{x}_i)\|_2^2
		+ \frac{1}{2}\sum\nolimits_{j=1}^N \|(W_{{r|X}}\Lambda^{\frac{1}{2}})^\top \phi_k(\bm{x}_j) \|_2^2 - \text{Tr}\left({W_e^\top W_r}\right),
	\end{array}
\end{equation}
where $\Lambda$ is automatically determined by the optimization.
KSVD optimization of Primal-Attention is easy to implement by adding a regularization loss. 
Hence, Primal-Attention not only represents self-attention with KSVD formulation in the primal, but utilizes the optimization of KSVD by regularizing the attention, satisfying the stationarity conditions of KSVD when reaching a {zero-value} objective.

\section{Numerical Experiments} 
\label{sec::exp}
In this section, we verify the effectiveness of our Primal-Attention applied in Transformers on {five well-established benchmarks: time series, long sequence modelling, reinforcement learning, image classification and language modelling.}
Notably, we consider two types of Transformers applied with our Primal-Attention, i.e., PrimalFormer (Primal.) and Primal.$+$.
\textit{i)} In PrimalFormer, Primal-Attention~\eqref{eq:outputs:Primal-Attention} is applied to all attention layers, regularized with the KSVD loss \eqref{eq:obj:std}. 
This setup is preferred when data shows relatively strong low-rank structure or the model redundancy is quite substantial.
\textit{ii)} Primal.$+$ refers to the baselines from the Transformer family with the last layer replaced by our Primal-Attention, which is in favour in large-scale data and complicated tasks where less information compression is desired in the learning, especially in shallower layers. 
To specify the Transformer backbone, we denote our method by ``Primal.$+$Backbone'' herein.
Primal.$+$ serves as a flexible variant combined with 
different Transformer backbones,
and with KSVD optimization applied through an implicit low-rank regularization loss, advocating
to learn more informative features,
as in Figure~\ref{fig:overview:low-rank}.
The two main hyper-parameters of our method are the 
coefficient $\eta$ in \eqref{eq::objective_function} and the number of projection directions $s$ of KSVD in \eqref{eq:ksvd:lssvm:std:attention}.
In data-dependent 
cases
with  $f(X)$, we 
take a subset of $X$ by uniformly sampling $n=\min\{s*10, N\}$ points from $X$ for efficiency aspects, {as the main patterns of a matrix can be retained with random linear projections shown by the Johnson–Lindenstrauss lemma~\cite{lindenstrauss1984extensions}.} 
Details are given in the Supplementary Material.

\paragraph{UEA Time Series Classification} 
\label{subsec::timeseries}
UEA Time Series Classification Archive~\cite{bagnall2018uea} is the benchmark for the evaluation on temporal sequences.
Following~\cite{wu2022flowformer}, we select 10 multivariate datasets with pre-processed data according to~\cite{zerveas2021transformer}, and employ 2-layer Transformer as backbone with the hidden dimension 512 on 8 heads and the embedding dimension 64 for self-attention.
{The hyper-parameter search of our method is with  $\eta\in\{0.1, 0.2, 0.5\}$, $s\in\{20, 30, 40\}$.} 
Experiments are run on one NVIDIA Tesla P100 SXM2 16GB GPU.

\begin{table}[t!]
	\caption{Test accuracy (\%) on  the UEA time series classification archive benchmark~\cite{bagnall2018uea} with comparisons to canonical Transformer (Trans.), Linear Transformer (Linear.), Reformer (Re.), Longformer (Long.), Performer (Per.), cosFormer (Cos.), Flowformer (Flow.), YOSO-E and SOFT.}
	\label{tab::timeseries}
	\begin{center}
		\resizebox{\textwidth}{!}{
			\begin{tabular}{cccccccccccc}
				\toprule
				\multirow{2}{*}{Dataset / Model} 
				& Trans. & Linear. & Re. & Long. & Per. & YOSO-E & Cos. & SOFT & Flow. & \multicolumn{2}{c}{\bf Ours} 
				\\ \cmidrule(lr){11-12}
				& \cite{vaswani2017attention} & \cite{katharopoulos2020transformers} & \cite{kitaev2020reformer} & \cite{beltagy2020longformer} & \cite{choromanski2021rethinking} & \cite{zeng2021you} & \cite{qin2022cosformer} & \cite{lu2021soft} & \cite{wu2022flowformer}
				& Primal. & Primal.$+$Trans. 
				\\ \midrule
				EthanolConcentration           
				& 32.7 & 31.9 & 31.9 & 32.3 & 31.2 & 31.2 & 32.3 & 33.5 & 33.8 
				& 33.1 & 35.4
				\\
				FaceDetection                  
				& 67.3 & 67.0 & 68.6 & 62.6 & 67.0 & 67.3 & 64.8 & 67.1 & 67.6      
				& 67.1 & 63.8
				\\
				HandWriting                    
				& 32.0 & 34.7 & 27.4 & 39.6 & 32.1 & 30.9 & 28.9 & 34.7 & 33.8       
				& 29.6 & 28.7            
				\\
				HeartBeat                      
				& 76.1 & 76.6 & 77.1 & 78.0 & 75.6 & 76.5 & 77.1 & 75.6 & 77.6
				& 76.1& 77.1             
				\\
				JapaneseVowels                 
				& 98.7 & 99.2 & 97.8 & 98.9 & 98.1 & 98.6 & 98.3 & 99.2 & 98.9
				& 98.4 & 98.9             
				\\
				PEMS-SF                        
				& 82.1 & 82.1 & 82.7 & 83.8 & 80.9 & 85.2 & 83.2 & 80.9 & 83.8   
				& 89.6 & 90.2              
				\\
				SelfRegulationSCP1             
				& 92.2 & 92.5 & 90.4 & 90.1 & 91.5 & 91.1 & 91.1 & 91.8 & 92.5
				& 92.5 & 92.5              
				\\
				SelfRegulationSCP2             
				& 53.9 & 56.7 & 56.7 & 55.6 & 56.7 & 53.9 & 55.0 & 55.6 & 56.1
				& 57.2 & 56.1              
				\\
				SpokenArabicDigits             
				& 98.4 & 98.0 & 97.0 & 94.4 & 98.4 & 98.9 & 98.4 & 98.8 & 98.8
				& 100 & 100       
				\\
				UWaveGestureLibrary            
				& 85.6 & 85.0 & 85.6 & 87.5 & 85.3 & 88.4 & 85.6 & 85.0 & 86.6
				& 86.3& 88.4              
				\\ \midrule
				Average Accuracy               
				& 71.9 & 72.4 & 71.5 & 72.0 & 71.9 & 72.2 & 71.5 & 72.2 & 73.0
				& 73.0 & \bf{73.1}               
				\\ \bottomrule
		\end{tabular}}
	\end{center}
	\vspace{-5mm}
\end{table}

\begin{table}[t!]
	\caption{Efficiency comparisons on running time and memory consumption on the UEA benchmark~\cite{bagnall2018uea} where running time (s/Epoch), the peak training memory usage (GB) are given. We report how much faster and how less memory each model uses than Transformer.}
	\label{tab::timeseries_efficiency}
	\begin{center}
		\resizebox{\textwidth}{!}{
				\begin{tabular}{ccccccccccccc}
					\toprule
					\multirow{2}{*}{\begin{tabular}[c]{@{}c@{}}Dataset /\\ seq. length\end{tabular}} & Ethanol. & Face. & Hand. & Heart. & Jap. & PEMS-SF & SCP1 & SCP2 & Arabic. & UWave & {\bf Avg.~Time} & {\bf Memory} 
					\\ 
					& 1751 & 62 & 152 & 405 & 26 & 144 & 896 & 1152 & 83 & 315 & {\bf (s/Epoch)} & {\bf (GB)}   
					\\ \midrule
					Trans.~\cite{vaswani2017attention}   
					& 4.3 (1$\times$) & 10.7 (1$\times$) & 0.3 (1$\times$) & 0.7 (1$\times$) & 0.5 (1$\times$) & 0.6 (1$\times$) & 1.8 (1$\times$) & 1.8 (1$\times$) & 3.7 (1$\times$) & 0.3 (1$\times$) & 2.5 (1$\times$) & 10.9 (1$\times$)
					\\
					Flow.~\cite{wu2022flowformer} 
					& 2.4 (1.8$\times$) & 9.8 (1.1$\times$) & 0.3 (1.0$\times$) & 0.7 (1.0$\times$) & 0.6 (0.8$\times$) & 0.7 (0.9$\times$) & {\bf 1.4 (1.3$\times$)} & {\bf 1.3 (1.4$\times$)} & 4.4 (0.8$\times$) & 0.3  (1.0$\times$) & 2.2 (1.1$\times$) & 2.8 (3.9$\times$)        
					\\ \cmidrule(lr){2-11} \cmidrule(lr){12-13}
					{\bf Primal.$+$Trans.}                                        
					& 3.3 (1.3$\times$) & {\bf 5.7 (1.9$\times$)} & 0.3 (1.0$\times$) & 0.7 (1.0$\times$) & 0.5 (1.0$\times$) & 0.7 (0.9$\times$) & 1.6 (1.1$\times$) & 1.6 (1.1$\times$) & 4.0  (0.9$\times$) & 0.3  (1.0$\times$) & {\bf 1.9 (1.3$\times$)} & 6.5 (1.7$\times$)       
					\\
					{\bf Primal.} 
					& {\bf 2.3 (1.9$\times$)} & 6.9 (1.6$\times$) & 0.4 (0.8$\times$) & {\bf 0.4 (1.8$\times$)} & 0.5 (1.0$\times$) & 0.8 (0.8$\times$) & {\bf 1.4 (1.3$\times$)} & {\bf 1.3 (1.4$\times$)} & 4.5 (0.8$\times$) & 0.4 (0.8$\times$) & {\bf 1.9 (1.3$\times$)} & {\bf 2.7 (4.0$\times$)}  
					\\ \bottomrule
			\end{tabular}}
	\end{center}
	\vspace{-5mm}
\end{table}
Table~\ref{tab::timeseries} reports the test accuracy of the compared recent methods, where the best result is in bold. 
Both our PrimalFormer and Primal.$+$ achieve comparable and better performance than the state-of-the-art results provided by Flowformer~\cite{wu2022flowformer}.
Notably, Primal.$+$ yields the best accuracy with $1.2\%$ overall improvement upon the canonical Transformer~\cite{vaswani2017attention}, i.e., by replacing the softmax-based self-attention with our Primal-Attention in the last layer.
This shows the promising potential of Primal-Attention in enhancing temporal modelling capacity upon the canonical softmax self-attention. 
It is worth mentioning that our PrimalFormer applying the KSVD optimization with low-rank regularization to all layers also obtains good performance. 
This could be due to the fact that these datasets are relatively simple where model redundancy can exist, so that appropriately imposing the low-rank regularization in KSVD does not harm the model expressiveness for the data. 
{We also compare the running time and memory consumption with the canonical Transformer and Flowformer, which is proposed recently with state-of-the-art performances and efficiency. 
	Both our Primal.~and Primal.$+$Trans.~consistently lead to improved efficiency than the canonical Transformer. 
	In Primal., all attention layers are implemented with the proposed Primal-Attention, while Primal.$+$Trans.~applies the Primal-Attention in the last layer, making Primal.~a more efficient model. 
	We note that our Primal.~further surpasses Flowformer in most cases regarding both running time and memory.}

\paragraph{Long-Range Arena Benchmark} 
\label{subsec::lra}
Long-Range Arena (LRA)~\cite{tay2020long} is a benchmark for the long-sequence scenarios, including equation calculation (ListOps)~\cite{nangia2018listops}, review classification (Text)~\cite{maas2011learning}, document retrieval (Retrieval)~\cite{radev2013acl}, image classification (Image)~\cite{krizhevsky2009learning} and image spatial dependencies (Pathfinder)~\cite{linsley2018learning}.
We follow the settings in~\cite{xiong2021nystromformer} with PyTorch.
The Transformer backbone is set with 2 layers, hidden dimension 128 with 2 heads and embedding dimension 64 with mean pooling, where Reformer (Re.) uses 2 hashes, Performer (Per.) has 256 random feature dimensions and Linformer (Lin.) uses a projection dimension of 256.
{Our hyper-parameter is set from $\eta\in\{0.05, 0.1\}$, $s\in\{20, 30\}$.}
Experiments are conducted on a single NVIDIA Tesla V100 SXM2 32GB GPU.

From the reported top-1 test accuracy in Table~\ref{tab::lra}, our PrimalFormer shows better accuracy while achieves top efficiency (see Table~\ref{tab::efficiency}) than several efficient self-attention counterparts including Reformer, Performer and Linformer.
Notably, our model Primal.$+$Trans.~achieves the state-of-the-art accuracy of $60.4\%$, which is $1.6\%$ higher than Transformer, and $1.2\%$ higher than the currently best YOSO-E, showing that the deployed Primal-Attention is able to boost the performance upon canonical Transformer distinctively. 
On top of it, Primal.$+$Trans.~has distinctively higher efficiency and require much less memory than the canonical Transformer as in Table~\ref{tab::efficiency}, {and our Primal.~outperforms all compared variants of efficient Transformer.} 
Compared to Table \ref{tab::timeseries} with simpler data, Primal.$+$Trans.~further achieves performance gain over PrimalFormer. 
The reason can be that the shallower (the first) layer needs more model capacity for depicting the data patterns in the learning, while the deep (the second) layer captures less detailed features and is regularized with more informative feature learning through our Primal-Attention.  
{The efficiency of our Primal-Attention are further pronounced with longer sequences than that of the UEA benchmark in Table \ref{tab::timeseries_efficiency}.}

\begin{table}[t]
	\caption{Test accuracy (\%) on the LRA benchmark~\cite{tay2020long} with comparisons to 
		canonical Transformer (Trans.), Reformer (Re.), Performer (Per.),
		Linformer (Lin.), Nystr\"{o}mformer (Nystr\"{o}m.), Longformer (Long.) and YOSO-E. 
	}
	\label{tab::lra}
	\begin{center}
		\resizebox{0.85\textwidth}{!}{
			\begin{tabular}{cccccccccc}
				\toprule
				\multirow{2}{*}{Dataset (seq.~length)} 
				& Trans. & Re. & Per. & Lin. & Nystr\"{o}m. & Long. & YOSO-E  &\multicolumn{2}{c}{\bf Ours} 
				\\ \cmidrule(lr){9-10}
				& \cite{vaswani2017attention} & \cite{kitaev2020reformer} & \cite{choromanski2021rethinking} & \cite{wang2020linformer} & \cite{xiong2021nystromformer} & \cite{beltagy2020longformer} & \cite{zeng2021you} & Primal. & Primal.$+$Trans.
				\\ \midrule
				ListOps (2K) 
				& 37.1 & 19.1 & 18.8 & 37.3 & 37.2 & 37.2 & 37.3 & 37.3 & 37.3                
				\\
				Text (4K)                            
				& 65.0 & 64.9 & 63.8 & 55.9 & 65.5 & 64.6 & 64.7 & 61.2 & 65.4                
				\\
				Retrieval (4K)                       
				& 79.4 & 78.6 & 78.6 & 79.4 & 79.6 & 81.0 & 81.2 & 77.8 & 81.0                
				\\
				Image (1K)  
				& 38.2 & 43.3 & 37.1 & 37.8 & 41.6 & 39.1 & 39.8 & 43.0 & 43.9               
				\\
				Pathfinder (1K)                     
				& 74.2 & 69.4 & 69.9 & 67.6 & 70.9 & 73.0 & 72.9 & 68.3 & 74.3                
				\\ \midrule
				Average Accuracy                 
				& 58.8 & 55.1 & 53.6 & 55.6 & 59.0 & 59.0 & 59.2 & 57.5 & \bf{60.4}           
				\\ \bottomrule
		\end{tabular}}
	\end{center}
	\vspace{-5mm}
\end{table}

\begin{table}[t]
	\caption{Efficiency comparisons on running time  and memory consumption on  LRA~\cite{tay2020long}.}
	\label{tab::efficiency}
	\begin{center}
		\resizebox{\textwidth}{!}{
			\begin{tabular}{ccccccccccc}
				\toprule
				\multirow{2}{*}{Model} & \multicolumn{5}{c}{Time (s/1K-steps)} & \multicolumn{5}{c}{Memory (GB)}     
				\\ \cmidrule(lr){2-6} \cmidrule(lr){7-11}
				& ListOps & Text & Retrieval & Image & Pathfinder & ListOps & Text & Retrieval & Image & Pathfinder 
				\\ \midrule
				Trans.~\cite{vaswani2017attention} 
				& 194.5 (1$\times$)& 694.8 (1$\times$)& 1333.7 (1$\times$)& 334.5 (1$\times$)& 405.5 (1$\times$) 
				& 5.50 (1$\times$)& 21.24 (1$\times$)& 18.72 (1$\times$)& 5.88 (1$\times$)& 5.88 (1$\times$)
				\\
				Nystr\"{o}m.~\cite{xiong2021nystromformer} 
				& 68.4 (2.8$\times$)& 120.9 (5.7$\times$)& 235.5 (5.7$\times$)& 179.5 (1.9$\times$)& 221.2 (1.8$\times$)
				& 0.89 (6.2$\times$)& 1.69 (12.6$\times$)& 3.29 (5.7$\times$)& 1.93 (3.0$\times$)& 1.93 (3.0$\times$)           
				\\
				Lin.~\cite{wang2020linformer} 
				& 63.4 (3.1$\times$)& 116.5 (6.0$\times$)& 226.2 (5.9$\times$)& 158.5 (2.1$\times$)& 204.0 (2.0$\times$)
				& 1.73 (3.2$\times$)& 3.45 (6.2$\times$)& 6.33 (3.0$\times$)& 3.45 (1.7$\times$)& 3.45 (1.7$\times$)         
				\\
				Per.~\cite{choromanski2021rethinking} 
				& 83.8 (2.3$\times$)& 157.5 (4.4$\times$)& 320.6 (4.2$\times$)& 211.4 (1.6$\times$)& 278.1 (1.5$\times$)
				& 1.67 (3.3$\times$)& 3.34 (6.4$\times$)& 6.28 (3.0$\times$)& 3.34 (1.8$\times$)& 3.34 (1.8$\times$)           
				\\
				Re.~\cite{kitaev2020reformer} 
				& 87.0 (2.2$\times$)& 168.5 (4.1$\times$)& 339.9 (3.9$\times$)& 223.7 (1.5$\times$)& 286.7 (1.4$\times$)
				& 1.64 (3.3$\times$)& 3.29 (6.5$\times$)& 6.09 (3.1$\times$)& 3.29 (1.8$\times$)& 3.29 (1.8$\times$)    
				\\ \cmidrule(lr){2-6} \cmidrule(lr){7-11}
				\bf{Primal.$+$Trans.} 
				& 113.4 (1.7$\times$) & 367.6 (1.9$\times$) & 546.5 (2.4$\times$) & 212.1 (1.6$\times$) & 263.2 (1.5$\times$)
				& 5.24 (1.1$\times$) & 20.7 (1.0$\times$) & 18.59 (1.0$\times$)& 5.35 (1.1$\times$) & 5.35 (1.1$\times$)
				\\
				\bf{Primal.} 
				& \bf{56.5 (3.4$\times$)}& \bf{93.6 (7.4$\times$)}& \bf{185.3 (7.2$\times$)}& \bf{142.9 (2.3$\times$)}& \bf{180.0 (2.3$\times$)}
				& \bf{0.69 (7.9$\times$)}& \bf{1.37 (15.5$\times$)}& \bf{2.99 (6.3$\times$)}& \bf{1.39 (4.2$\times$)}& \bf{1.52 (3.9$\times$)}   
				\\ \bottomrule
		\end{tabular}}
	\end{center}
	\vspace{-5mm}
\end{table}

\paragraph{Reinforcement Learning}
\label{subsec::reinforcement}
We consider the offline reinforcement learning (RL) performance of our methods on D4RL benchmark~\cite{fu2020d4rl} designed for continuous control tasks.
We choose three different environments controlling the robot movement: HalfCheetah, Hopper and Walker.
Our experiments are conducted on three datasets pre-collected under different policies: Medium-Expert, Medium and Medium-Replay.
We follow the experimental settings in Flowformer~\cite{wu2022flowformer}, and also compare to Decision Transformer (DT)~\cite{chen2021decision} which is commonly considered as the baseline with state-of-the-art performances based on the canonical attention.
Note that offline RL is an auto-regressive task and our Primal-Attention is adapted to its causal version, as described in the Supplementary Material.
We adopt the architecture of 3 layers, hidden dimension 256 with 4 heads, and the embedding dimension 64. 
Our hyper-parameters are set as {$\eta=0.05$, $s \in \{32, 64, 96\}$.}
Each experiment is run with 3 different seeds on one NVIDIA Tesla P100 SXM2 16GB GPU. 

Results in Table~\ref{tab::rl} demonstrate that our Primal.$+$ outperforms all compared methods with a distinctly higher average reward.
Specifically, our Primal.$+$DT reaches $4.0$ points higher than the state-of-the-art Flowformer \cite{wu2022flowformer}. 
Compared to the baseline Decision Transformer (DT), our Primal.$+$ only replaces the self-attention in the top layer and keeps other structures the same and manages to improve the average reward by $5.3$ points.
This verifies the effectiveness of our Primal-Attention and the benefits of our low-rank regularization of KSVD for the generality of DT in offline reinforcement learning. 
{We evaluate the efficiency with comparisons to DT and Flowformer in Table~\ref{tab::rl:efficiency}, which give distinctively better results than other baselines in Table~\ref{tab::rl}. 
	Our Primal.$+$DT achieves comparable time and memory efficiency as the most efficient baseline DT, while Flowformer shows significantly lower efficiency in both aspects. Recall that our Primal.$+$DT achieves an average reward of 77.5, which is 5.3 points higher than DT and 4.0 points higher than Flowformer.}

\begin{table}[t]
\caption{Rewards on D4RL~\cite{fu2020d4rl} datasets. 
	We report mean and variance for 3 seeds.
	A higher reward and a lower deviation indicates better performance.
	We consider Decision Transformer (DT), Linear Transformer (Linear.), Reformer (Re.), Performer (Per.), cosFormer (Cos.) and Flowformer (Flow.).
}
\label{tab::rl}
\begin{center}
	\resizebox{\textwidth}{!}{
		\begin{tabular}{ccccccccc}
			\toprule
			\multirow{2}{*}{Dataset}       
			& \multirow{2}{*}{Environment} & DT & Linear. & Re. & Per. & Cos. & Flow. & \bf{Ours}      
			\\ 
			& & \cite{chen2021decision} & \cite{katharopoulos2020transformers} & \cite{kitaev2020reformer} & \cite{choromanski2021rethinking} & \cite{qin2022cosformer} & \cite{wu2022flowformer} & Primal.$+$DT 
			\\ \midrule
			\multirow{3}{*}{\begin{tabular}[c]{@{}c@{}}Medium\\ -Expert\end{tabular}} 
			& HalfCheetah                  
			& 83.8$\pm$3.3 & 78.2$\pm$3.2 & 81.5$\pm$1.6 & 85.1$\pm$2.1 & 85.5$\pm$2.9 & 90.8$\pm$0.4 & {77.8$\pm$22.1}
			\\
			& Hopper                       
			& 104.0$\pm$2.5 & 107.2$\pm$0.9 & 104.2$\pm$9.8 & 93.5$\pm$13.9 & 98.1$\pm$7.4 & 109.9$\pm$1.0 & 111.5$\pm$0.2 
			\\
			& Walker                       
			& 107.7$\pm$0.6 & 67.2$\pm$27.3 & 71.4$\pm$1.8 & 72.6$\pm$2.4 & 100.5$\pm$14.5 & 108.0$\pm$0.4 & 108.9$\pm$0.1            
			\\ \cmidrule(lr){3-9}
			\multirow{3}{*}{Medium}        
			& HalfCheetah                 
			& 42.4$\pm$0.1 & 42.3$\pm$0.2 & 42.2$\pm$0.1 & 42.1$\pm$0.2 & 42.1$\pm$0.3 & 42.2$\pm$0.2 & 43.0$\pm$0.0
			\\
			& Hopper                      
			& 64.2$\pm$1.1 & 58.7$\pm$0.4 & 59.9$\pm$0.7 & 59.7$\pm$7.5 & 59.8$\pm$3.8 & 66.9$\pm$2.5 & 74.5$\pm$0.6          
			\\
			& Walker                       
			& 70.6$\pm$3.2 & 57.9$\pm$10.6 & 65.8$\pm$4.9 & 63.3$\pm$10.7 & 71.4$\pm$1.2 & 71.7$\pm$2.5 & 77.9$\pm$7.8 
			\\ \cmidrule(lr){3-9}
			\multirow{3}{*}{\begin{tabular}[c]{@{}c@{}}Medium\\ -Replay\end{tabular}} 
			& HalfCheetah                  
			& 34.6$\pm$0.6 & 32.1$\pm$1.5 & 33.6$\pm$0.7 & 31.7$\pm$0.9 & 32.8$\pm$3.6 & 34.7$\pm$1.5 & 38.9$\pm$0.4            
			\\
			& Hopper                       
			& 79.7$\pm$7.4 & 74.3$\pm$7.0 & 66.1$\pm$2.6 & 64.6$\pm$24.2 & 59.3$\pm$16.5 & 75.5$\pm$14.5 & 88.5$\pm$12.5
			\\
			& Walker                       
			& 62.9$\pm$5.0 & 62.1$\pm$7.4 & 50.1$\pm$3.5 & 61.3$\pm$6.7 & 60.5$\pm$9.9 & 62.0$\pm$3.1 & 76.8$\pm$10.3           
			\\ \midrule
			\multicolumn{2}{c}{Average Reward}                           
			& 72.2$\pm${\bf 2.6} & 64.4$\pm$6.5 & 63.9$\pm$2.9 & 63.8$\pm$7.6 & 67.8$\pm$7.6 & 73.5$\pm$2.9 & {\bf 77.5}$\pm$6.0         
			\\ \bottomrule
	\end{tabular}}
\end{center}
\vspace{-5mm}
\end{table}

\begin{table}[t]
\caption{{Efficiency comparisons on running time (s/1K-steps) and memory (GB) on D4RL~\cite{fu2020d4rl}.
	}
}
\label{tab::rl:efficiency}
\begin{center}
	\resizebox{0.85\textwidth}{!}{
			\begin{tabular}{ccccccccccc}
				\toprule
				\multirow{2}{*}{Model} & \multicolumn{2}{c}{Medium-Expert} & & \multicolumn{2}{c}{Medium} & & \multicolumn{2}{c}{Medium-Replay} \\
				\cmidrule{2-3}  \cmidrule{5-6} \cmidrule{8-9} 
				& Time & Memory &   & Time & Memory &   & Time & Memory
				\\ \midrule
				DT (average reward: 72.2) &  20.8 & 0.3 & & 20.8 &0.3 & &20.8 & 0.3 
				\\
				Flow. (average reward: 73.5) & 54.4& 1.5 & & 54.4&1.5& & 54.3& 1.5  \\
				Primal.$+$DT  (average reward: {77.5}) & 23.5 & 0.3& & 23.4&0.3& & 23.3& 0.3 
				&
				\\ \bottomrule
		\end{tabular}}
\end{center}
\vspace{-5mm}
\end{table}

\paragraph{{Large-scale Experiments}}
\label{subsec::imagenet}
We evaluate the capability of our Primal.$+$ model with DeiT-Small/16~\cite{touvron2021training} as backbone on 
ImageNet-100~\cite{russakovsky2015imagenet} and ImageNet-1K~\cite{deng2009imagenet} for image classification task. 
We also experiment with the language modelling task on WikiText-103~\cite{merity2016pointer}. 
On ImageNet,
we train DeiT-Small/16 and our Primal.$+$DeiT-Small/16 from scratch following the same training protocols in~\cite{touvron2021training} with 4 NVIDIA Tesla V100 SXM2 32GB GPUs. 
Our hyper-parameters are chosen as $\eta=0.05$, $s \in \{32, 64, 96\}$. 
{On WikiText-103, we follow the setting in~\cite{peng2021random} where the sequence length is set as 512, the model consists of 6 decoder layers with 8 heads.
We implement the causal version of Primal-Attention in the last layer, i.e., Primal.$+$Trans with $\eta=0.1$, $s=30$.
Models are trained from scratch on 4 NVIDIA Tesla V100 SXM2 32GB GPUs for
150K updates after 6K-steps warm-up.}

{Table~\ref{tab::large_scale}(a) provides the test accuracy, training time and memory on a single V100 GPU with batch size 256 on both ImageNet-100 and ImageNet-1K datasets.
There is only one set of time and memory since models follow the same training protocols on both datasets.
Our Primal.$+$ achieves better accuracy than DeiT-Small/16 on ImageNet-100.
It also achieves the same accuracy as baseline with less training time and memory on ImageNet-1K. 
On WikiText-103 in Table~\ref{tab::large_scale}(b), our method with default setups significantly reduces the perplexity by 2.0 points than Transformer, and achieves comparable performances with the well-tuned Flowformer, a latest SoTA model, with enhanced efficiency.
Results show that using Primal-Attention in the last layer helps the overall performance.}

\begin{table}[t]
\centering
\renewcommand{\arraystretch}{1.1}
\caption{{Results on large-scale experiments including image classification and language modelling.}}
	\begin{minipage}{0.6\textwidth}
		\centering
		\scalebox{0.66}
		{\subfloat[Test accuracy (\%) and efficiency on ImageNet-100~\cite{russakovsky2015imagenet} and ImageNet-1K~\cite{deng2009imagenet}.]{
				\begin{tabular}{ccccc}
					\toprule
					Model & \begin{tabular}[c]{@{}c@{}}ImageNet-100 \\ (Top-1 Acc.)\end{tabular} & \begin{tabular}[c]{@{}c@{}}ImageNet-1K \\ (Top-1 Acc.)\end{tabular} & \begin{tabular}[c]{@{}c@{}}Time \\ (s/1K-steps)\end{tabular} & \begin{tabular}[c]{@{}c@{}}Memory \\ (GB)\end{tabular}  
					\\ \midrule
					DeiT-Small/16           
					& 74.2 & 79.8 & 2425.5 & 14.2          
					\\
					Primal.$+$DeiT-Small/16 
					& \textbf{75.7} & 79.8 & \textbf{2330.2} & \textbf{14.0}
					\\ \bottomrule
				\end{tabular}
				\label{tab::imagenet}}}
	\end{minipage}\hfill
	\begin{minipage}{0.4\textwidth}
		\centering
		\scalebox{0.66}
		{\subfloat[Results on WikiText-103~\cite{merity2016pointer}.]{
				\begin{tabular}{cccc}
					\toprule
					\multirow{2}{*}{Model} & \multirow{2}{*}{\begin{tabular}[c]{@{}c@{}}Perplexity\\ ($\downarrow$)\end{tabular}} & \multirow{2}{*}{\begin{tabular}[c]{@{}c@{}}Time\\ (s/1K-steps)\end{tabular}} & \multirow{2}{*}{\begin{tabular}[c]{@{}c@{}}Memory\\ (GB)\end{tabular}} 
					\\ 
					& & &  
					\\ \midrule
					Trans.~\cite{vaswani2017attention}                 
					& 33.0 & 3108.4 & 9.0                                   
					\\
					Flow.~\cite{wu2022flowformer}
					& \textbf{30.8} & 3998.4 & 10.5
					\\
					Primal.$+$Trans.       
					& 31.0 & \textbf{3104.0} & \textbf{8.9}      
					\\ \bottomrule
				\end{tabular}
				\label{tab::wikitext}}}
	\end{minipage}\hfill
\label{tab::large_scale}
\vspace{-3mm}
\end{table}

\paragraph{Ablation Study on Using Two Projections $e(\bm x)$, $r(\bm x)$}
We conduct an ablation study with (w/) and without (w/o) projection scores, i.e., $r$-scores, involving the left singular vectors, on LRA~\cite{tay2020long}. 
Table \ref{tab::lra:r:scores} shows that using both projections (w/ $r$-scores) helps boost performances, verifying the effectiveness of learning with the two sides of singular vectors on an asymmetric attention kernel.

\begin{table}[t]
\caption{{Ablation on $r$-scores involving left singular vectors on LRA~~\cite{tay2020long} with test accuracy (\%).} 
}
\label{tab::lra:r:scores}
\begin{center}
	\resizebox{0.7\textwidth}{!}{
			\begin{tabular}{cccccccc}
				\toprule
				Model & $r$-scores & ListOps & Text & Retrieval & Image & Pathfinder & Avg.~Acc 
				\\ \midrule
				\multirow{2}{*}{Primal.}          
				& w/o 
				& 36.8 & 52.4 & 58.2 & 30.5 & 50.2 & 45.6   
				\\
				& w/  
				& \bf 37.3 & \bf 61.2 & \bf 77.8 & \bf 43.0	& \bf 68.3 & \bf 57.5
				\\ \cmidrule{3-8}
				\multirow{2}{*}{Primal.$+$Trans.} 
				& w/o 
				& 37.1 & 65.1 & 79.2 & 42.8 & 72.8 & 59.4   
				\\
				& w/  
				& \bf 37.3 & \bf 65.4 & \bf 81.0 & \bf 43.9 & \bf 74.3 & \bf 60.4 
				\\ \bottomrule
		\end{tabular}}
\end{center}
\vspace{-5mm}
\end{table}

\section{Related work}
\label{sec::related_work}

{Since the pioneering work~\cite{tsai2019}, the kernel-based approaches have become popular in Transformers, in which the kernel interpretation on the attention matrix has been shed light on.}
FourierFormer~\cite{nguyen2022fourierformer} treats the canonical self-attention as non-parametric regression with {methodologies for} symmetric kernels.
\cite{chi2022kerple} considers relative positional embedding with conditional positive definite kernel.
\cite{nguyen2023a} treats self-attention operation as support vector regression {without considering the asymmetry in the deployed kernel methods, and the  supervised regression is not applied in optimzing the attention either}.
\cite{chen2021skyformer} addresses the issue of asymmetry, however, it {resorts to symmetrization by replacing the softmax attention with an approximated symmetric one}, {thereby still dismissing the asymmetry}. 
{These prior works deploy} kernel-based techniques {that are originally designed for symmetric kernels and request to suffice Mercer conditions}, which is inconsistent with the asymmetric {nature} in self-attention, {resulting in a nontrivial gap between the analytical understanding and numerical implementation towards revealing the rationale in Transformers}. 
{In \cite{wright2021transformers}, it leverages the kernel tricks from RKBS~\cite{zhang2009reproducing} that allows asymmetry and formulates attention} as a binary kernel learning problem via empirical risk minimization.
However, it is hard to find an explicit optimization accordingly in {implementing Transformers.} 
Nevertheless, \cite{wright2021transformers} {provides an analytical tool from the aspect of kernel representer theorem upon RKBS that allows asymmetry.}

Much literature has also devoted to improving the efficiency of the attention computation through different approximation techniques. 
In the related works addressing the attention mechanism,
Reformer~\cite{kitaev2020reformer} uses locally-sensitive hashing for sparse approximation. 
Performer~\cite{choromanski2021rethinking} approximates self-attention matrix with random features. Linformer~\cite{wang2020linformer} considers low-rank approximation with the help of random projections. 
Nystr\"{o}mformer~\cite{xiong2021nystromformer} utilizes the Nystr\"{o}m method by down sampling the queries and keys in the attention matrix. \cite{child2019generating} incorporates sparsity prior on attention. 
These works pose the focus on  reducing the computation of the attention kernel matrix from the canonical self-attention. 
Hence, these works all address how to solve the problem in the dual form involving the kernel matrix, while we work in a significantly different way, that is, in the primal form.

\section{Conclusion}
\label{sec::conclusion}
In this paper, we interpret the self-attention in Transformers with asymmetric kernels and construct a learning framework with SVD on asymmetric kernels (KSVD) under the setups of LSSVM. 
Within the context of KSVD, a primal-dual model representation is formulated for self-attention and 
a novel attention mechanism (Primal-Attention) is proposed accordingly by leveraging the primal representation. 
It is quite significant that with Primal-Attention, not only the computation of the attention kernel matrix in the dual can be avoided, but also the cast unsupervised KSVD optimization can be efficiently incorporated into the training  through an additional regularization loss for more informative low-rank property. The analytical derivations and numerical evaluations demonstrate our great potentials in bridging explicit model interpretability and state-of-the-art performances.
Future works can include developing different variants with the low-rank property, e.g., robust Transformers, investigating more general applications of Primal-Attention to a wide range of architectures and tasks.

\section*{Acknowledgements}
This work is jointly supported by the European Research Council under the European Union’s Horizon 2020 research and innovation program/ERC Advanced Grant E-DUALITY (787960), iBOF project Tensor Tools for Taming the Curse (3E221427), Research Council KU Leuven: Optimization framework for deep kernel machines C14/18/068,  KU Leuven Grant CoE PFV/10/002, The Research Foundation–Flanders (FWO) projects: GOA4917N (Deep Restricted kernel Machines: Methods and Foundations), Ph.D./Postdoctoral grant, the Flemish Government (AI Research Program), EU H2020 ICT-48 Network TAILOR (Foundations of Trustworthy AI-Integrating Reasoning, Learning and Optimization), Leuven.AI Institute.

\bibliography{bibfile}
\bibliographystyle{unsrt} 

\appendix
~\\
\section*{Appendix}

In this material, we present the proofs of all analytical results in the paper and additional comments in Section~\ref{sec::proofs}.
More experimental details and results are also provided in Section~\ref{sec::more_exp}.
We also provide the broader impacts of our work in Section~\ref{sec::broader_impact}.

\section{Theoretical Proofs}
\label{sec::proofs}
In this section, we provide the proofs of all analytical results presented in the paper, covering Theorem~\ref{theorem:ksvd:dual}, Remark~\ref{rmk:primal:dual}, and Lemma~\ref{prop:obj:0}. 
Additional comments are also provided following each proof in this material.

\subsection{Proof of Theorem 3.2}

\begin{proof}[Proof of Theorem~\ref{theorem:ksvd:dual}]
Given the matrix $X \in \mathbb R^{N\times d}$ consists of sequence data $\{\bm x_{i} \in \mathbb{R}^d\}_{i=1}^N$, the primal optimization problem in self-attention of KSVD with the constructed data-dependent projection weights  is formulated as follows, i.e.,~\eqref{eq:ksvd:lssvm:std:attention} in the paper:
\begin{equation}\label{eq::primal_fx}
	\begin{aligned}
		\max\limits_{W_e, W_r, \bm e_i, \bm r_j }
		&  \, 
		J=\frac{1}{2}\sum\nolimits_{i=1}^N \bm e_i^\top\Lambda \bm e_i + \dfrac{1}{2}\sum\nolimits_{j=1}^N  \bm r_j^\top \Lambda \bm r_j
		- \text{Tr}\left(W_e^\top W_r \right) 
		\\
		{\rm s.t.}\quad
		&  \bm{e}_i = (f(X)^\top W_e)^\top \phi_q (\bm x_i)  , \,\,  i=1, \ldots, N,\\
		& \bm{r}_j =  (f(X)^\top W_r)^\top  \phi_k (\bm x_j), \,\, j=1, \ldots, N,
	\end{aligned}
\end{equation}
where the projection weights of the feature maps $\phi_q(\cdot), \phi_k(\cdot): \mathbb{R}^d \to \mathbb{R}^p$ can be further denoted as $f(X)^\top W_e=:W_{e|X}\in\mathbb{R}^{p\times s}$, $f(X)^\top W_r=:W_{r|X}\in\mathbb{R}^{p\times s}$ relying on parameters
$W_e, W_r \in \mathbb R^{N\times s}$ and the constant transformation matrix $f(X)=: F_X \in \mathbb R^{N\times p}$, the regularization coefficient denoted by $\Lambda \in \mathbb{R}^{s\times s}$ is a positive diagonal matrix.

The Lagrangian of \eqref{eq::primal_fx} is
\begin{equation}
	\label{eq:lagrange:obj:data:case3}
	\begin{aligned}
		& \mathcal{L}(W_e, W_r, \bm{e}_i, \bm{r}_j, \bm h_{e_i}, \bm h_{r_j})  
		= \frac{1}{2}\sum\nolimits_{i=1}^N \bm e_i^\top\Lambda\bm e_i + \dfrac{1}{2}\sum\nolimits_{j=1}^N  \bm r_j^\top \Lambda \bm r_j 
		-  \text{Tr}\left(W_e^\top W_r\right)   
		\\
		& - \sum\nolimits_{i=1}^N \bm h_{e_i}^\top \left(\bm e_i - W_e^\top f(X) \phi_q(\bm x_i)\right) - \sum\nolimits_{j=1}^N \bm h_{r_j}^\top \left(\bm r_j - W_r^\top f(X) \phi_k(\bm x_j)\right),
	\end{aligned}
\end{equation}
where two sets of dual variable vectors, i.e., $\bm{h}_{e_i}, \bm{h}_{r_j} \in \mathbb R^s$, are introduced to the equality constraints regarding the projection scores $\bm{e}_i$ and $\bm{r}_j$, for $i,j=1,\ldots, N$, respectively.

By taking the partial derivatives to the Lagrangian~\eqref{eq:lagrange:obj:data:case3}, the Karush-Kuhn-Tucker (KKT) conditions then lead to:
\begin{equation}\label{eq:lssvm:kkt:nonlinear:data:case3}
	\left\{
	\begin{aligned}
		\frac{\partial \mathcal{L}}{\partial W_e} = 0 \implies & W_r = \sum\nolimits_{i=1}^N  f(X)\phi_q(\bm x_i)  \bm h_{e_i}^\top,  
		\\
		\frac{\partial \mathcal{L}}{\partial W_r} = 0 \implies & W_e =\sum\nolimits_{j=1}^N f(X) \phi_k(\bm x_j) \bm h_{r_j}^\top 
		\\ 
		\frac{\partial \mathcal{L}}{\partial \bm e_i} = 0 \implies & {\Lambda}\bm e_i =  \bm h_{e_i},\,  i=1,\ldots,N,
		\\
		\frac{\partial \mathcal{L}}{\partial \bm r_j} = 0 \implies & {\Lambda} \bm r_j = \bm h_{r_j}, \,  j=1,\ldots,N,
		\\
		\frac{\partial \mathcal{L}}{\partial \bm h_{e_i}} = 0 \implies & W_e^\top f(X) \phi_q(\bm x_i) = \bm e_i, \,  i=1,\ldots,N,
		\\
		\frac{\partial \mathcal{L}}{\partial \bm h_{r_j}} = 0 \implies & W_r^\top f(X) \phi_k(\bm x_j) = \bm r_j, \,  i=1,\ldots,N.
	\end{aligned}
	\right.
\end{equation}

By eliminating the primal variables $W_e$, $W_r$ in KKT conditions~\eqref{eq:lssvm:kkt:nonlinear:data:case3}, we then have
\begin{equation*}
	\left\{
	\begin{aligned}
		& \sum\nolimits_{j=1}^N  \bm h_{r_j} \phi_k(\bm x_j)^\top f(X)^\top f(X) \phi_q(\bm x_i) = \Lambda^{-1}  \bm h_{e_i}, \, i = 1,\ldots,N,
		\\
		& \sum\nolimits_{i=1}^N   \bm h_{e_i} \phi_q(\bm x_i)^\top f(X)^\top  f(X) \phi_k(\bm x_j) = \Lambda^{-1}  \bm h_{r_j}, \, j = 1,\ldots,N,
	\end{aligned}
	\right.
\end{equation*}
which can be expressed in the matrix form as
\begin{equation*}
	\left[
	\begin{array}{cc}
		\bm 0_{N\times N} & \left[  \phi_q(\bm x_i)^\top f(X)^\top f(X) \phi_k(\bm x_j)\right]
		\\
		\left[  \phi_k(\bm x_j)^\top f(X)^\top  f(X) \phi_q(\bm x_i)\right]   
		& \bm 0_{N\times N}
	\end{array}
	\right] 
	\left[
	\begin{array}{c}
		H_e  
		\\
		H_r 
	\end{array}
	\right]
	= 
	\left[
	\begin{array}{c}
		H_e  
		\\ 
		H_r 
	\end{array}
	\right]
	\Lambda^{-1},
\end{equation*}
with $H_e = [\bm h_{e_1}, \ldots, \bm h_{e_N}]^\top \in \mathbb R^{N\times s}$ and $H_r = [\bm h_{r_1}, \ldots, \bm h_{r_N}]^\top \in \mathbb R^{N\times s}$. 

Therefore, the optimization problem of KSVD in the dual yields the following shifted eigenvalue problem with an asymmetric kernel matrix $K$, such that:
\begin{equation} \label{eq::shifted_eigen_fx}
	\begin{aligned}
		K H_r& = H_e \Sigma,
		\\
		K^\top H_e& = H_r \Sigma,
	\end{aligned}
\end{equation}
which collects the solutions corresponding to the non-zero entries in $\Lambda$ such that 
$\Sigma \triangleq \Lambda^{-1}$. 
The asymmetric kernel $K$ contains the entries induced as
$K_{ij} := \left< f(X)\phi_q(\bm x_i), f(X)\phi_k(\bm x_j) \right> =: \left< \phi_q'(\bm x_i), \phi_k'(\bm x_j) \right>$, $i,j=1,\ldots,N$. 
From the Lanczos Decomposition Theorem~\cite{lanczos1958linear}, i.e., Theorem 2.2 in the paper, we can see that the solutions to the dual problem of KSVD in self-attention, i.e., $H_e$ and $H_r$, correspond to the left and right singular vectors of the asymmetric kernel matrix $K$, 
where $\Sigma$ serves as  the corresponding  singular values. 
This completes the proof.

\end{proof}

\paragraph{Comments on Theorem 3.2} 
With the primal problem in~\eqref{eq:ksvd:lssvm:std:attention} in the paper, Theorem~\ref{theorem:ksvd:dual} provides the corresponding dual problem of KSVD formulated for self-attention. 
In~\cite{suykens2016svd}, a novel variational principle is proposed for SVD with LSSVMs, where a primal-dual formulation for the matrix (linear) SVD is derived. 
Our KSVD leverages the kernel-based learning framework from~\cite{suykens2016svd}, however, in addition to our specific application of interpreting self-attention, there are other significant differences and non-trivial novelties in our work: 
\begin{itemize}
\item[\emph{i)}]~\cite{suykens2016svd} mainly addresses the original SVD given any data matrix, while we formulate the non-linear extension leading to the asymmetric attention matrix in relation to the queries and keys. 
Additionally,~\cite{suykens2016svd} presents the optimization w.r.t.~a single projection direction in the linear SVD, while we generalize the formulation to multiple projection directions in the matrix form. 

\item[\emph{ii)}] The data sources for the two non-linear feature maps are related to the queries and keys, as opposed to~\cite{suykens2016svd} that specifies the two data sources as the rows and columns of the given data matrix. 
Therefore, our KSVD is more general in the data setups. 

\item[\emph{iii)}] Rather than using only the data-independent projection weights $W_e, W_r$ as linear mappings in~\cite{suykens2016svd}, we propose the generalized form that allows extra transformation matrix dependent non-linearly on the sequence data for self-attention.
\end{itemize}

In particular, the benefits and motivations of our data-dependent projection weights are as follows: 
\begin{itemize}
\item[\emph{i)}] In the canonical self-attention, the values vary for different input sequence data, and later in Remark~\ref{rmk:primal:dual}, we show that the values can be regarded as playing the role of the dual variables in KSVD. Inspired by this property, we introduce input sequence data information to the corresponding primal variables.

\item[\emph{ii)}] In the proposed Primal-Attention, the data-dependent projection weights provide more degrees of freedom to improve model's representation ability.

\item[\emph{iii)}] Using data-dependent projection weights does not affect the derivation of the shifted eigenvalue problem in the dual.
Specifically, when the transformation matrix $F_X$ is chosen as an identity matrix for a simpler structure, it boils downs to the data-independent case, where the  kernel $K$ in self-attention is obtained with entries $K_{ij}=\langle \phi_q(\bm x_i), \phi_k(\bm x_j) \rangle$, $i,j=1,\ldots,N$.
\end{itemize}
Provided with the general form of the projections weights in \eqref{eq::primal_fx}, practitioners can flexibly adjust the KSVD setups for the self-attention implementation.  
Related empirical studies can be referred to Section~\ref{subsec::data_dependent} in this material.

\subsection{Proof of Remark~3.3}
With the derivations of the primal-dual optimization problems above, the primal-dual model representation of our KSVD problem can be provided correspondingly.
The proof of Remark~\ref{rmk:primal:dual} in the paper closely follows the proof of Theorem~\ref{theorem:ksvd:dual}, and we show it as follows.
\begin{proof}[Proof of Remark~\ref{rmk:primal:dual}]
The primal model representations for the self-attention outputs
in~\eqref{eq::primal_fx} are
\begin{equation}\label{eq::primal_representation}
	\begin{aligned}
		\text{Primal:} &\quad 
		\left\{
		\begin{array}{l}
			{e}(\bm{x}) =  (f(X)^\top W_e)^\top \phi_q (\bm{x}),
			\\
			{r}(\bm{x}) =  (f(X)^\top W_r)^\top \phi_k (\bm{x}).
		\end{array}
		\right.     
	\end{aligned}
\end{equation}
The dual model representations for the self-attention outputs can be derived by eliminating the primal variables with~\eqref{eq:lssvm:kkt:nonlinear:data:case3}:
\begin{equation}\label{eq::dual_representation}
	\begin{aligned}
		\text{Dual:} &\quad 
		\left\{
		\begin{array}{l}
			\begin{aligned}
				{e}(\bm{x}) = W_e^\top f(X)\phi_q(\bm{x})
				& = \big(\sum\nolimits_{j=1}^N f(X)\phi_k(\bm{x}_j)\bm{h}_{r_j}^\top \big)^\top f(X)\phi_q(\bm{x})
				\\
				& = \sum\nolimits_{j=1}^N   \bm  h_{r_j} \phi_q(\bm{x})^\top  f(X)^\top  f(X) \phi_k(\bm{x_j}),
			\end{aligned}
			\\
			\begin{aligned}
				{r}(\bm{x}) = W_r^\top f(X)\phi_k(\bm{x})
				& = \big(\sum\nolimits_{i=1}^N f(X)\phi_q(\bm{x}_i)\bm{h}_{e_i}^\top \big)^\top f(X)\phi_k(\bm{x})
				\\
				& = \sum\nolimits_{i=1}^N  \bm  h_{e_i}    \phi_q(\bm{x_i})^\top f(X)^\top f(X) \phi_k(\bm{x}). 
			\end{aligned}
		\end{array}
		\right.
	\end{aligned}
\end{equation}
Further, with the kernel trick in the dual optimization problem \eqref{eq::shifted_eigen_fx}, i.e.,
\begin{equation*}
	\kappa(\bm x_i, \bm x_j) := \left<  f(X) \phi_q(\bm x_i), f(X)\phi_k(\bm x_j) \right>, \, i,j=1,\ldots,N,
\end{equation*}
we then attain the primal-dual representations of KSVD that allows data-dependent projection weights for self-attention as follows:
\begin{equation*}
	\begin{array}{rll}
		\text{Primal:} & 
		\left\{
		\begin{array}{l}
			{e}(\bm{x}) = W_{e|X}^\top  \phi_q(\bm{x})
			\\
			{r}(\bm{x}) = W_{r|X}^\top   \phi_k(\bm{x})
		\end{array},
		\right.
		\quad
		\text{Dual:} &
		\left\{
		\begin{array}{l}
			{e}(\bm{x})  = \sum\nolimits_{j=1}^N   \bm  h_{r_j}    \kappa(\bm{x},\bm{x}_j) 
			\\
			{r}(\bm{x}) = \sum\nolimits_{i=1}^N  \bm   h_{e_i}
			\kappa(\bm{x}_i,\bm{x})
		\end{array}
		\right.
	\end{array},
\end{equation*}
where $W_{e|X}:=f(X)^\top W_e\in\mathbb{R}^{p\times s}$, $W_{r|X}:=f(X)^\top W_r\in\mathbb{R}^{p\times s}$.
\end{proof}

\paragraph{Comments on Remark 3.3}
With Remark~\ref{rmk:primal:dual}, we can equivalently represent the projection scores in different ways, i.e., either through the feature maps in the primal or the kernel matrix in the dual. 
Under the framework of KSVD, the existing attention outputs can be interpreted as the projection scores $e(\bm x)$ in the dual representation, where the values correspond to the dual variables $\bm h_{r_j}$. 
The primal-dual models provide versatile alternatives to represent and understand the attention outputs. 
Notably, the primal representation can avoid the computation of the kernel matrix which is widely considered as an obstacle to the computational efficiency of attention.
In addition, we find that there exists another set of projections reflecting the asymmetry information, i.e., $r(\bm x)$. 
Motivated by the above, we propose our new self-attention mechanism, i.e., Primal-Attention in Section~\ref{sec::method} in the paper.

\subsection{Proof of Lemma 4.2}
Lemma~\ref{prop:obj:0} evaluates the objective value $J$ in the primal optimization problem \eqref{eq::primal_fx} when the solutions satisfy the stationarity conditions in \eqref{eq::shifted_eigen_fx}. 
\begin{proof}[Proof of Lemma~\ref{prop:obj:0}]
Based on the KKT conditions in~\eqref{eq:lssvm:kkt:nonlinear:data:case3}, by eliminating the primal variables $W_e, W_r, \bm{e}_i, \bm{r}_j$, the optimization objective $J$ is given by

\begin{equation}\label{eq::kkt_fx}
	\begin{aligned}
		J  & =  \dfrac{1}{2}\sum\nolimits_{i=1}^N \bm{e}_i^\top\Lambda\bm{e}_i 
		+ \dfrac{1}{2}\sum\nolimits_{j=1}^N  \bm r_j^\top \Lambda \bm r_j - \text{Tr}\left(W_e^\top W_r\right) 
		\\
		& =  \dfrac{1}{2}\sum\nolimits_{i=1}^N \bm{e}_i^\top\Lambda\bm{e}_i 
		+ \dfrac{1}{2}\sum\nolimits_{j=1}^N  \bm r_j^\top \Lambda \bm r_j - \text{Tr}\left(W_r^\top W_e\right) 
		\\
		& =  \dfrac{1}{2} \sum\nolimits_{i=1}^N  ({\Lambda^{-1}} \bm{h}_{e_i})^\top \Lambda \Lambda^{-1} \bm h_{e_i} 
		+ \dfrac{1}{2} \sum\nolimits_{j=1}^N  ({\Lambda^{-1}}\bm{h}_{r_j})^\top \Lambda \Lambda^{-1} \bm h_{r_j}
		\\
		&  \quad \quad \quad - \text{Tr}\bigg( \Big(\sum\nolimits_{i=1}^N f(X)\phi_q(\bm{x}_i)  \bm{h}_{e_i}^\top \Big)^\top    
		\sum\nolimits_{j=1}^N  f(X) \phi_k(\bm{x}_j) \bm{h}_{r_j}^\top \bigg)
		\\
		& =  \dfrac{1}{2} \sum\nolimits_{i=1}^N \bm{h}_{e_i}^\top \Lambda^{-1} \bm{h}_{e_i} 
		+ \dfrac{1}{2} \sum\nolimits_{j=1}^N  \bm{h}_{r_j}^\top \Lambda^{-1} \bm{h}_{r_j} 
		\\
		&  \quad \quad \quad - \text{Tr}\bigg( \Big(\sum\nolimits_{i=1}^N f(X)\phi_q(\bm{x}_i)  \bm{h}_{e_i}^\top \Big)^\top    
		\sum\nolimits_{j=1}^N  f(X) \phi_k(\bm{x}_j) \bm{h}_{r_j}^\top \bigg)
		\\
		& =  \dfrac{1}{2} \sum\nolimits_{i=1}^N  \bm h_{e_i}^\top \Lambda^{-1} \bm h_{e_i} 
		+ \dfrac{1}{2} \sum\nolimits_{j=1}^N  \bm h_{r_j}^\top \Lambda^{-1} \bm h_{r_j}
		- \text{Tr}\left( H_e^\top K H_r \right)
		\\
		& \overset{\eqref{eq::shifted_eigen_fx}}{=}  
		\dfrac{1}{2} \sum\nolimits_{i=1}^N  \bm h_{e_i}^\top \Sigma \bm h_{e_i} + \dfrac{1}{2} \sum\nolimits_{j=1}^N  \bm h_{r_j}^\top \Sigma \bm h_{r_j} 
		- \text{Tr}\left( H_e^\top H_e \Sigma\right)
		\\
		& = \dfrac{1}{2} \sum\nolimits_{l=1}^s {\sigma}_l  \bm h_{e, l}^\top  \bm h_{e,l} + \dfrac{1}{2} \sum\nolimits_{l=1}^s {\sigma}_l  \bm h_{r, l}^\top \bm h_{r,l}
		- \sum\nolimits_{l=1}^s {\sigma}_l \bm h_{e,l}^\top \bm h_{e,l},
	\end{aligned}
\end{equation}
where in the last equation, we denote the dual variables corresponding to the $l$-th projection direction, i.e., singular vectors in relation to the singular value $\sigma_l$,  as $ \bm h_{e,l} := H_e[:, l] = [\bm h_{e_1}[l],  \ldots, \bm h_{e_N}[l]]^\top  \in \mathbb R^{N}$,  
$\bm h_{r,l} := H_r[:, l] = [\bm h_{r_1}[l], \ldots, \bm h_{r_N}[l]]^\top  \in \mathbb R^{N}$, and $\Sigma = \text{diag}\{\sigma_1, \ldots, \sigma_s\} \triangleq \Lambda^{-1}$. 

Based on both~\eqref{shifted:eigen:ksvd:std:attention} and Theorem~\ref{theorem:ksvd:dual} in the paper, we have the following equations:
\begin{equation*}\label{eq:refer:obj:0}
	\begin{aligned}
		& K^\top K \bm{h}_{r,l}  = \sigma_l K^\top \bm{h}_{e,l} = \sigma_l^2 \bm{h}_{r,l},
		\\
		& K K^\top \bm{h}_{e,l}  = \sigma_l K \bm{h}_{r,l} = \sigma_l^2  \bm{h}_{e,l}.
	\end{aligned}
\end{equation*}
Hence, we can rewrite
\begin{equation} \label{eq::equal}
	\begin{aligned}
		\bm{h}_{e,l}^\top \bm h_{e, l}
		& = \frac{1}{\sigma_l^2}(K K^\top \bm{h}_{e,l})^\top \bm{h}_{e,l}
		= \frac{1}{\sigma_l^2} \bm{h}_{e,l}^\top K K^\top \bm{h}_{e,l}
		= \frac{1}{\sigma_l^2} \bm{h}_{e,l}^\top (K K^\top \bm{h}_{e,l})
		\\
		& = \frac{1}{\sigma_l^2} \bm{h}_{e,l}^\top (\sigma_l K \bm{h}_{r,l})
		= \frac{1}{\sigma_l} \bm{h}_{e,l}^\top K \bm{h}_{r,l}
		= \frac{1}{\sigma_l} (K^\top \bm{h}_{e,l} )^\top \bm{h}_{r,l}
		\\ 
		& = \bm{h}_{r,l}^\top \bm{h}_{r,l},
	\end{aligned}
\end{equation}
which leads to 
\begin{equation}\label{eq::zero_obj}
	\begin{aligned}
		J  
		&  = \frac{1}{2} \sum\nolimits_{l=1}^s {\sigma_l}  \bm h_{e,l}^\top  \bm h_{e,l} + \frac{1}{2} \sum\nolimits_{l=1}^s {\sigma_l}  \bm h_{r, l}^\top \bm h_{r,l} - \sum\nolimits_{l=1}^s {\sigma_l} \bm h_{e,l}^\top \bm h_{e,l}
		\\
		& \overset{\eqref{eq::equal}}{=} \frac{1}{2} \sum\nolimits_{l=1}^s {\sigma_l}  \bm h_{e, l}^\top  \bm h_{e,l} + \dfrac{1}{2} \sum\nolimits_{l=1}^s {\sigma_l}  \bm h_{e, l}^\top  \bm h_{e,l}
		- \sum\nolimits_{l=1}^s {\sigma_l} \bm h_{e,l}^\top \bm h_{e,l}
		\\
		& = 0.
	\end{aligned}
\end{equation}
Note that \eqref{eq::kkt_fx}, \eqref{eq::equal} and \eqref{eq::zero_obj} also hold for the data-independent projection weights case where $f(X)$ is an identity matrix.
In this case, the entries in the induced asymmetric kernel $K$ become $K_{ij}= \langle f(X)\phi_q(\bm x_i),  f(X)\phi_k(\bm x_j)\rangle
=\langle \phi_{q}(\bm{x}_i), \phi_{k}(\bm{x}_j)\rangle$,
$i,j=1,\ldots,N$.
This completes the proof.
\end{proof} 

\paragraph{Comments on Lemma 4.2}
With Lemma~\ref{prop:obj:0}, we validate that the objective $J$ \eqref{eq::primal_fx} in the primal optimization problem reaches zero when the stationarity conditions are satisfied, i.e., the singular vectors and their corresponding singular values in \eqref{eq::shifted_eigen_fx} are obtained. 
In the paper, the KSVD optimization for self-attention is realized by incorporating the objective $J$ as an additional regularization loss to the original task-oriented loss, and then minimizing the total loss to zero as shown in~\eqref{eq::objective_function} and~\eqref{eq:obj:std} in the paper.
In this manner, we avoid solving the dual optimization that involves a SVD problem on a kernel matrix.
Moreover, as in the proof of Theorem~\ref{theorem:ksvd:dual}, we note that the regularization coefficient $\Lambda$ in the primal optimization~\eqref{eq::primal_fx} corresponds to the singular values in the dual optimization~\eqref{eq::shifted_eigen_fx}.
With the SGD-based or AdamW-based algorithm, we flexibly integrate the hyper-parameter selection of $\Lambda$ into the optimization by setting $\Lambda$ as a learnable parameter.
In this case, $\Lambda$ can be optimized together with other model parameters by simply minimizing the total loss in~\eqref{eq::objective_function} in the paper.

\section{More Experimental Results}
\label{sec::more_exp}

\subsection{Setup Details}
This section provides the implementation details of all experiments included in the paper.
Firstly, we outline the main algorithm of our Primal-Attention mechanism in Algorithm~\ref{alg::primal} for clarity.
{Note that in data-dependent cases with $f(X)$, we take a subset of
$X$ by uniformly sampling $n$ points from $X$ for efficiency aspects, as the main patterns of a matrix can be retained with random linear projections shown by the Johnson–Lindenstrauss lemma~\cite{lindenstrauss1984extensions}. 
This will be illustrated in details in the following.}

\begin{algorithm}[ht]
\caption{Learning with Primal-Attention }
\label{alg::primal}
\begin{algorithmic}
	\Require 
	$X:=[{\bm{x}}_1,\ldots,{\bm{x}}_N]^\top \in\mathbb{R}^{N\times d}$ is the input sequence to the attention block in Transformer, 
	mappings $g_q(\cdot): \mathbb{R}^{d_q}\to\mathbb{R}^p$, $g_k(\cdot):\mathbb{R}^{d_k}\to\mathbb{R}^p$ defined in~\eqref{eq:ksvd:lssvm:std:attention} in the paper, 
	number of projection directions $s$ defined in~\eqref{eq:ksvd:lssvm:std:attention} in the paper,
	regularization coefficient $\eta$ defined in~\eqref{eq::objective_function} in the paper,
	\Ensure
	Transformation matrix $f(X)=:F_X \in\mathbb{R}^{N\times p}$ defined in~\eqref{eq:ksvd:lssvm:std:attention} in the paper is required if data-dependent projection weights are used.
	\If{Data-dependent projection weights}
	\State $q({\bm x}_i)=W_q{\bm x}_i$, $k({\bm x}_i)=W_k{\bm x}_i$;
	\Comment{$W_q \in \mathbb{R}^{d_q \times d}$, 
		$ W_k \in \mathbb{R}^{d_k \times d}$}
	\State $e({\bm x}_i) = (f(X)^\top W_e)^\top g_q(q({\bm x}_i))$; 
	\Comment{compute $e$-score for $i=1,\ldots,N$}
	\State $r({\bm x}_i) = (f(X)^\top W_r)^\top g_k(k({\bm x}_i))$; 
	\Comment{compute $r$-score for $i=1,\ldots,N$}
	\State ${\bm o}_i = W_o[e({\bm x}_i); r({\bm x}_i)]$;
	\Comment{compute concatenated output with $W_o\in\mathbb{R}^{ d_v \times(2s)}$}
	\ElsIf{Data-independent projection weights}
	\State $q({\bm x}_i)=W_q{\bm x}_i$, $k({\bm x}_i)=W_k{\bm x}_i$;
	\Comment{$W_q \in \mathbb{R}^{d_q \times d}$, 
		$ W_k \in \mathbb{R}^{d_k \times d}$}
	\State $e({\bm x}_i) = W_e^\top g_q(q({\bm x}_i))$; 
	\Comment{$W_e \in\mathbb{R}^{p\times s}$, compute $e$-score for $i=1,\ldots,N$}
	\State $r({\bm x}_i) = W_r^\top g_k(k({\bm x}_i))$; 
	\Comment{$W_r \in\mathbb{R}^{p\times s}$, compute $r$-score for $i=1,\ldots,N$}
	\State ${\bm o}_i = W_o[e({\bm x}_i); r({\bm x}_i)]$;
	\Comment{compute concatenated output with $W_o\in\mathbb{R}^{ d_v \times(2s)}$}
	\EndIf
\end{algorithmic}
\end{algorithm}

\paragraph{UEA Time Series}
The UEA time series benchmark~\cite{bagnall2018uea} consists of 30 datasets.
Following the setup in~\cite{wu2022flowformer}, we select 10 datasets for evaluation.
For all experiments of our PrimalFormer and Primal.$+$Trans., we adopt the data-dependent projection weights for Primal-Attention, i.e., we have $W_{e|X}:=f(X)^\top W_e \in \mathbb{R}^{p\times s}$ 
and $W_{r|X}:=f(X)^\top W_r \in \mathbb{R}^{p\times s}$.
On account that some datasets consist of long sequence samples, e.g., EthanolConcentration of length 1751,
SelfRegulationSCP1 of length 896,
SelfRegulationSCP2 of length 1152,
our choice of $f(X)$ should include more information about $X$ for greater model flexibility while maintaining computational efficiency.
In this regard, we set $f(X):=X'$ where $X'\in \mathbb{R}^{n \times p}$ is a subset of the sequence data $X \in \mathbb R^{N\times d}$ by uniformly sampling $n=\min\{s*{\tt rank\_multi}, N\}$ points (rows) from $X$.
We set ${\tt rank\_multi}=10$ for most cases, and set ${\tt rank\_multi}=5$ for datasets including FaceDetection, HandWriting, JapaneseVowels, PEMS-SF and SpokenArabicDigits, since they have shorter sequence lengths.
In this manner, the size of the transformation matrix $f(X)$ is implemented as $\mathbb{R}^{n \times p}$ with $n\ll N$, reducing memory requirements especially for long sequence data with large $N$.

\paragraph{Long-Range Arena}
Long-Range Arena (LRA)~\cite{tay2020long} consists of long-sequence scenarios: ListOps of 2K sequence length, Text of 4K length, Retrieval of 4K, Image of 1K and Pathfinder of 1K.
With joint consideration of performance and efficiency, for all experiments of our PrimalFormer and Primal.$+$Trans.~in the paper, we adopt Primal-Attention with data-dependent projection weights and set $n=\min\{s*10, N\}$, i.e., ${\tt rank\_multi}=10$, for all cases.

\begin{figure}[t]
\centering    
{\includegraphics[width=\textwidth]{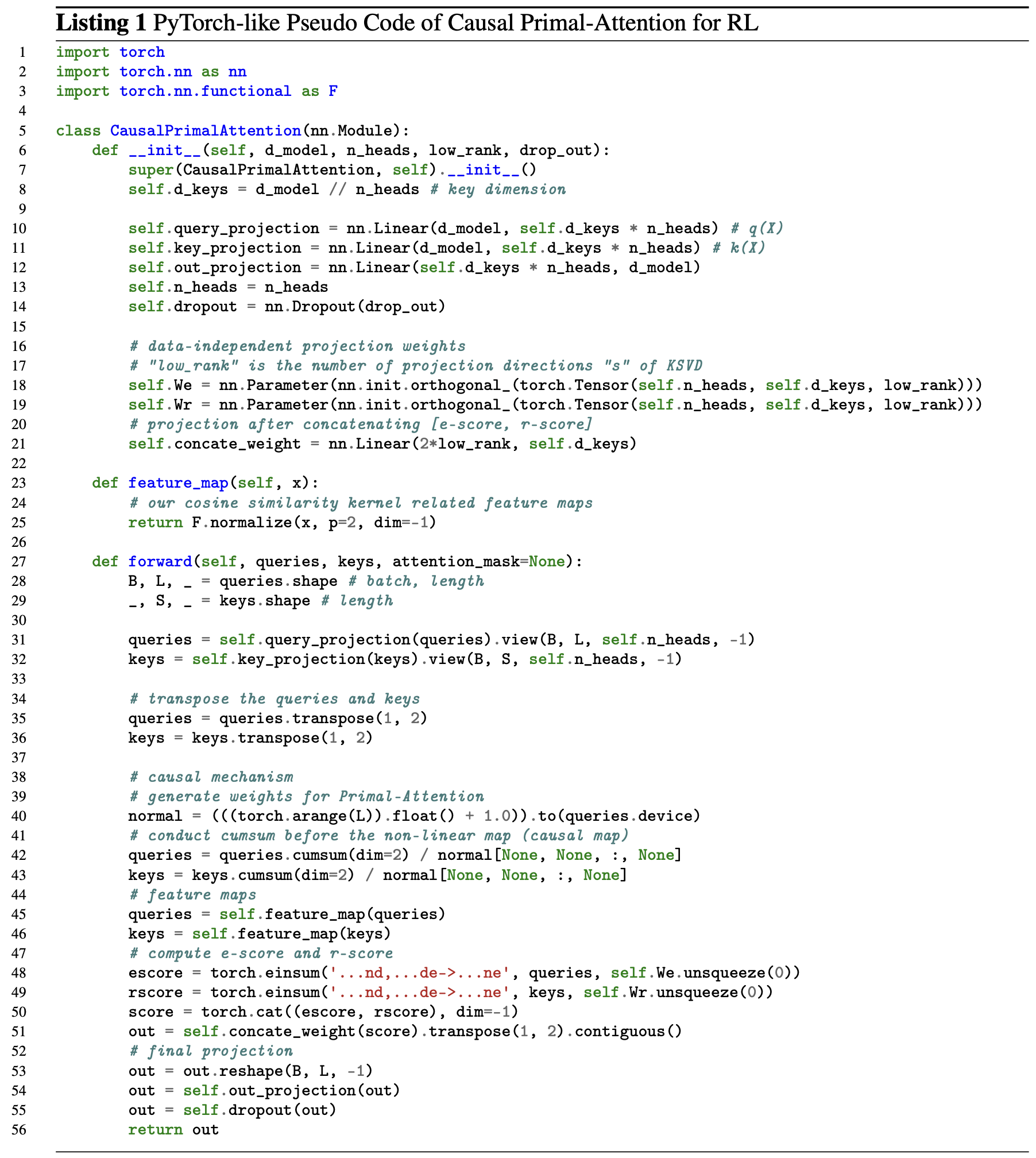}}
\caption{PyTorch-like pseudo code of Causal Primal-Attention for RL.
}
\label{listing::causal}
\end{figure}

\paragraph{Reinforcement Learning}
D4RL~\cite{fu2020d4rl} is a suite of continuous control tasks and datasets for benchmarking progress in offline reinforcement learning.
In this experiment, we consider Primal.$+$DT with Decision Transformer (DT)~\cite{chen2021decision} as the backbone.
Specifically, we consider a three-layer DT with its self-attention in third layer replaced by our Primal-Attention.
As DT utilizes a causal self-attention mask which predicts actions autoregressively, to align with the causal structure, we propose the causal-version of Primal-Attention, namely, Causal Primal-Attention. 
For clarity, we attach the corresponding PyTorch-like pseudo code in Figure~\ref{listing::causal} in this material.
Note that for this task, we utilize Causal Primal-Attention with the simpler data-independent projection weights, i.e., $W_e, W_r\in\mathbb{R}^{p\times s}$, which helps to prevent overfitting in learning rewards in the RL training process.

\paragraph{Image Classification}
ImageNet-100~\cite{russakovsky2015imagenet} contains 100 classes of images
from ImageNet-1K~\cite{deng2009imagenet}.
On both ImageNet-100 and ImageNet-1K,
we use Primal.$+$DeiT-Small/16 with standard DeiT-Small/16 as the backbone.
Specifically, the self-attention of the last layer of DeiT-Small/16 is replaced by our Primal-Attention using data-dependent projection weights with the setup $n=\min\{s*10, N\}$, i.e.,~${\tt rank\_multi}=10$.

\paragraph{Language Modelling} We conduct the language modelling on the WikiText-103~\cite{merity2016pointer}, which aims to estimate the probability distribution of a token given the previous ones.
We replace the self-attention in the last layer of the Transformer baseline with our Causal Primal-Attention using data-dependent projection weights with the setup $n=\min\{s*10, N\}$, i.e.,~${\tt rank\_multi}=10$.

\subsection{Further Ablation Studies}

\begin{table}[t]
\caption{Ablation study on the two main hyper-parameters $\eta$ and $s$. 
	We report test accuracy (\%) of PrimalFormer on the UEA time series classification archive benchmark~\cite{bagnall2018uea}.}
\label{tab::ablation_timeseries}
\begin{center}
	\resizebox{\textwidth}{!}{
		\begin{tabular}{cccccccccccc}
			\toprule
			\multirow{2}{*}{Dataset} & \multirow{2}{*}{$s$} & \multicolumn{4}{c}{$\eta$} 
			& \multirow{2}{*}{Dataset} & \multirow{2}{*}{$s$} & \multicolumn{4}{c}{$\eta$} 
			\\ \cmidrule(lr){3-6} \cmidrule(lr){9-12}
			& & 0  & 0.1  & 0.2  & 0.5  & 
			& & 0  & 0.1  & 0.2  & 0.5  
			\\ \midrule
			\multirow{3}{*}{EthanolConcentration} 
			& 20 & 32.3 &  31.6 & 30.8 & 32.7     
			& \multirow{3}{*}{FaceDetection}
			& 20 & 64.2 & \bf 67.1 & 66.2 & 66.4     
			\\ 
			& 30 & 30.0 & \bf 33.1 & 31.9 & 30.4 &                          
			& 30 & 65.0 & 65.3 & 65.2 & 65.7   
			\\
			& 40 & 30.8 & 32.3 & \bf 33.1 & 31.9 &                          
			& 40 & 64.5 & 65.6 & 66.5 & 66.7  
			\\ \midrule
			\multirow{3}{*}{HandWriting}        
			& 20 & 26.7 & 28.4 & 27.3 & 26.9     
			& \multirow{3}{*}{HeartBeat}        
			& 20 & 75.1 & 72.7 & 74.2 & 75.1
			\\
			& 30 & 28.2 & 26.9 & \bf 29.6 & 25.9 &                          
			& 30 & 75.6 & 75.6 & 75.6 & \bf 76.1     
			\\
			& 40 & 26.0 & 26.5 & 27.7 & 27.5 &                          
			& 40 & \bf 76.1 & 75.1 & 72.2 & 74.6 
			\\ \midrule
			\multirow{3}{*}{JapaneseVowels}        
			& 20 & 98.1 & 98.1 & 97.3 & 97.8    
			& \multirow{3}{*}{PEMS-SF}     
			& 20 & 86.1 & 85.6 & 83.8 & 84.4
			\\
			& 30 & 98.1 & 97.6 & 97.3 & 97.6 &                          
			& 30 & 83.8 & 86.7 & 82.1 & 86.7     
			\\
			& 40 & 98.1 & 98.1 & 97.6 & \bf 98.4 &                            
			& 40 & 86.1 & \bf 89.6 & 86.7 & 85.0      
			\\ \midrule
			\multirow{3}{*}{SelfRegulationSCP1}     
			& 20 & 91.5 & \bf 92.5 & 90.8 & 91.8    
			& \multirow{3}{*}{SelfRegulationSCP2}
			& 20 & \bf 57.2 & 53.9 & 55.6 & 56.1     
			\\
			& 30 & 92.2 & 91.8 & 92.2 & \bf 92.5 &                          
			& 30 & 55.6 & 53.9 & 55.6 & 55.8     
			\\
			& 40 & 91.8 & 91.8 & 91.5 & 91.8 &                          
			& 40 & 52.9 & 56.1 & 53.9 & 53.3     
			\\ \midrule
			\multirow{3}{*}{SpokenArabicDigits}   
			& 20 & 100 & 100 & 100 & 100      
			& \multirow{3}{*}{UWaveGestureLibrary} 
			& 20 & \bf 86.3 & 83.8 & 83.8 & 84.7     
			\\
			& 30 & 100 & 100 & 100 & 100 &                          
			& 30 & 85.3 & 85.0 & 85.0 & 84.1     
			\\
			& 40 & 100 & 100 & 100 & 100 &                          
			& 40 & \bf 86.3 & 84.1 & 84.1 & 85.9     
			\\ \bottomrule
	\end{tabular}}
\end{center}
\vspace{-5mm}
\end{table}

\begin{table}[t]
\caption{Ablation study on hyper-parameter $s$ with $\eta=0.05$. 
	We report the rewards of Primal.$+$DT on the D4RL datasets~\cite{fu2020d4rl}.
	A higher reward indicates better performance.}
\label{tab::ablation_rl}
\begin{center}
	\resizebox{0.77\textwidth}{!}{
		\begin{tabular}{cccccccccc}
			\toprule
			\multirow{2}{*}{\begin{tabular}[c]{@{}c@{}}Dataset / \\ Environment \end{tabular}} 
			& \multicolumn{3}{c}{Medium-Expert} 
			& \multicolumn{3}{c}{Medium} 
			& \multicolumn{3}{c}{Medium-Replay} 
			\\ \cmidrule(lr){2-4} \cmidrule(lr){5-7} \cmidrule(lr){8-10}
			& $s$=32 & $s$=64 & $s$=96 & $s$=32 & $s$=64 & $s$=96 & $s$=32 & $s$=64      & $s$=96      
			\\ \cmidrule(lr){2-4} \cmidrule(lr){5-7} \cmidrule(lr){8-10}
			HalfCheetah                                                               
			& 56.9 & 73.1 & \bf 75.6 
			& 42.9 & \bf 43.1 & 42.8 
			& \bf 39.5 & 37.9 & 39.3      
			\\
			Hopper                                                                    
			& 111.2 & \bf 112.0 & 111.1 
			& 66.7 & 63.1 & \bf 73.8 
			& 84.2 & \bf 91.7 & 87.3      
			\\
			Walker   
			& 108.7 & 108.9 & \bf 109.0 
			& 75.1 & \bf 77.1 & 77.0 
			& 71.8 & 70.4 & \bf 80.5      
			\\ \bottomrule
	\end{tabular}}
\end{center}
\vspace{-5mm}
\end{table}

\paragraph{Ablation on $\eta$ and $s$}
The numerical investigations are conducted on the two main hyper-parameters of our Primal-Attention, i.e., the coefficient $\eta$ of the KSVD regularization loss and the number of projection directions $s$.
We consider the UEA time series datasets.
The results of PrimalFormer, i.e., two-layer Transformer with Primal-Attentions, are given in Table~\ref{tab::ablation_timeseries} in this material.
\emph{Firstly}, compared to $\eta=0$, a rough tuning of $\eta > 0$ improves the performance for most of  the datasets.
For example, $\eta > 0$ on FaceDetection leads to consistent improvement over $\eta=0$.
This indicates that the KSVD optimization through the regularization loss $J$ in~\eqref{eq:obj:std} in the paper indeed brings performance boost over its non-regularized counterpart.
\emph{Secondly}, even without the KSVD optimization, i.e., $\eta=0$, our Primal-Attention already leads to good performance, such as the results on SpokenArabicDigits, SelfRegulationSCP2 with $s=20$, and UWaveGestureLibrary with $s=20, 40$. 
This verifies that the new representation in Primal-Attention in~\eqref{eq:primal:dual:ksvd:std:attention} in the paper can effectively represent the self-attention and conduct effective learning in the attention outputs.
\emph{Thirdly}, effective learning features can be captured in less dimensions than the original embedding dimension and a performance boost can be potentially gained with even fewer dimensions through our formulated KSVD.
To be specific, the embedding dimension for each head is 64, and we set $s\in\{20, 30, 40\}$ in the experiments.
Recall that the average accuracy of the canonical Transformer is 71.9\% in Table~\ref{tab::timeseries} in the paper, while our PrimalFormer reaches 73.1\%, which is 1.2\% higher upon the canonical one.
These results hence show that an appropriate compression in the number of projection directions by KSVD could lead to performance improvements when the low-rank property is desired.
Note that since the  kernel matrix in the dual of our Primal-Attention is of size $K \in \mathbb{R}^{N\times N}$, we limit $s$ up to $N$, i.e., $0 < s \leq N$, as there exists at most $N$ projection directions in the corresponding KSVD.
In general, larger $s$ is preferred in more complicated tasks with more sophisticated dependency between samples in the sequence data.
For instance, the reward learning in RL is such a case where less information compression is desired.  
This can be verified by the results given in Table~\ref{tab::ablation_rl} in this material, where the best performance is attained with $s$ as 64 or 96 in almost all cases.

\paragraph{Projection Weights}
\label{subsec::data_dependent}

\begin{table}[t]
\caption{Ablation on data-dependent and data-independent projection weights of our Primal-Attention mechanism. 
	We report the test accuracy (\%) of Primal.$+$Trans.~on UEA time series datasets~\cite{bagnall2018uea}.}
\label{tab::ablation_weights_time}
\begin{center}
	\resizebox{\textwidth}{!}{
		\begin{tabular}{cccccccccccc}
			\toprule
			\multirow{3}{*}{\begin{tabular}[c]{@{}c@{}}Data- \\ dependent \end{tabular}} & \multicolumn{10}{c}{Dataset}        
			& \multirow{3}{*}{\begin{tabular}[c]{@{}c@{}}Avg. \\ Acc.\end{tabular} } 
			\\ \cmidrule(lr){2-11}
			& \begin{tabular}[c]{@{}c@{}}Ethanol \\ Concen. \end{tabular} 
			& \begin{tabular}[c]{@{}c@{}}Face \\ Detec. \end{tabular} 
			& \begin{tabular}[c]{@{}c@{}}Hand \\ Writ. \end{tabular}  
			& \begin{tabular}[c]{@{}c@{}}Heart \\ Beat \end{tabular}  
			& \begin{tabular}[c]{@{}c@{}}JPN \\ Vowels \end{tabular}  
			& \begin{tabular}[c]{@{}c@{}}PEMS \\ -SF \end{tabular}                  
			& \begin{tabular}[c]{@{}c@{}}SelfRegu. \\ SCP1 \end{tabular} 
			& \begin{tabular}[c]{@{}c@{}}SelfRegu. \\ SCP2 \end{tabular}
			& \begin{tabular}[c]{@{}c@{}}Spoken \\ ArabicDig. \end{tabular} 
			& \begin{tabular}[c]{@{}c@{}}UWave \\ GestureLib. \end{tabular} 
			\\ \midrule
			No     
			& 34.6 & 63.5 & \bf 29.3 & 76.1 & \bf 99.2 & 88.4 & \bf 92.8 & \bf 58.3 & 100 & 87.2   
			& 72.9
			\\
			Yes                   
			& \bf 35.4 & \bf 63.8 & 28.7 & \bf 77.1 & 98.9 & \bf 90.2 & 92.5 & 56.1 & 100 & \bf 88.4
			& \bf 73.1
			\\ \bottomrule
	\end{tabular}}
\end{center}
\vspace{-5mm}
\end{table}

\begin{table}[t]
\caption{Ablation on data-dependent and data-independent projection weights of our Primal-Attention mechanism. 
	We report the test accuracy (\%) of Primal.$+$Trans.~on the LRA benchmark~\cite{tay2020long}.}
\label{tab::ablation_weights_lra}
\begin{center}
	\resizebox{0.77\textwidth}{!}{
		\begin{tabular}{ccccccc}
			\toprule
			\multirow{2}{*}{\begin{tabular}[c]{@{}c@{}}Data- \\ dependent \end{tabular}} & \multicolumn{5}{c}{Dataset}        
			& \multirow{2}{*}{Average Accuracy} 
			\\ \cmidrule(lr){2-6}
			& ListOps & Text & Retrieval & Image & Pathfinder &                           \\ \midrule
			No                   
			& 37.0 & 40.2 & 74.3 & 80.8 & \bf 65.6 & 59.6                           \\
			Yes                      
			& \bf 37.3 & \bf 43.9 & 74.3 & \bf 81.0 & 65.4 & \bf 60.4      
			\\ \bottomrule
	\end{tabular}}
\end{center}
\vspace{-5mm}
\end{table}

We investigate the effects of projection weights in the data-dependent and data-independent cases for Primal-Attention.
Tables~\ref{tab::ablation_weights_time} and~\ref{tab::ablation_weights_lra} in this material present the comparisons  between data-dependent  and data-independent projection weights used in  Primal.$+$Trans.~on UEA time series datasets~\cite{bagnall2018uea} and also LRA benchmark~\cite{tay2020long}.
On both benchmarks, data-dependent projection weights case surpasses its data-independent counterpart.
The reason of these results can be that data-dependent projection weights help increasing the model's representation ability and capturing more informative features from the rather long sequences in these datasets.
Furthermore, for the data-dependent case, we set $f(X):=X'$ where $X' \in \mathbb{R}^{n \times p}$ is a subset of sequence data by uniformly sampling  $n=\min\{s*{\tt rank\_multi}, N\}$ points from $X \in \mathbb{R}^{N \times p}$. 
As shown in Table \ref{tab::ablation_timeseries} in this material, for a given ${\tt rank\_multi}$ in each dataset, the increase of $s$ does not make the results fluctuate much. 
Similar phenomenon is found on the LRA datasets during our experiments.
Therefore, for almost all experiments in the paper, we simply set $n=\min\{s*10, N\}$ as default.
This can also serve as a mild suggestion for practitioners in implementation.
We note that data-dependent projection weights are not always in favor.
For example, in the RL tasks, model is prone to overfit the learning of rewards during training if we adopt the Primal-Attention with data-dependent projection weights.
Hence, we take the data-independent case instead.
In the generalized form of the projection weights with Primal-Attention, more possibilities of greater model representation ability are provided to fit various tasks and datasets.

\subsection{Further Remarks on Efficiency}
\paragraph{Efficiency with Primal-Attention in architectures} From our learning scheme in Figure~\ref{fig::pipeline} in the paper and the empirical efficiency analysis 
with Tables~\ref{tab::timeseries_efficiency}, \ref{tab::efficiency}, \ref{tab::rl:efficiency} and \ref{tab::large_scale} in the paper, we can see that the efficiency gain of the Transformers implemented with Primal-Attention 
over canonical baselines is influenced by two main factors,
\textit{i)} the number of Primal-Attention layers employed in the architecture, i.e., the more the better; 
\textit{ii)} sequence length of the training data, i.e., the longer the more significant:
\begin{itemize}
\item[\emph{i)}] With deep architectures, the efficiency can be further improved by replacing more layers with our Primal-Attention. 
Yet, in very deep Transformers, Primal-Attention is not necessarily always superior in performance when being applied to all layers, as the learning in shallow layers may not enjoy the benefits from the low-rank property from KSVD as much as the higher layers do. 
It would be interesting to explore a more generic implementation setup for Primal-Attention in very deep Transformers, as briefly mentioned in the last paragraph of possible future work in this material.
\item[\emph{ii)}] The length of the data sequence, i.e., $N$, is also a key factor influencing the efficiency. 
By avoiding the computation of the $N\times N$ attention matrix, our Primal-Attention can gain better efficiency on longer-sequence datasets. 
Although ImageNet-1K is large-scale, currently Transformers treat each image as a sequence of length 197 (with {\tt cls} token), which is actually not too long (even compared to some UEA datasets as shown in Table~\ref{tab::timeseries_efficiency} in the paper). 
Hence, this is also a reason why our Primal.$+$DeiT-Small/16 does not improve the efficiency significantly in Table~\ref{tab::large_scale}(a) in the paper. 
Similarly in WikiText-103, the data sequence length is 512, which is also not really long, hence the efficiency of our Primal.$+$Trans.~is not always superior under the current setups.
\end{itemize}
\paragraph{Efficiency gain of Primal.$+$ in different Tasks}
The efficiency gain of Primal.$+$ over baseline is more significant on UEA and LRA, as the backbone has only 2 layers, hence replacing one layer makes a difference to the overall architecture. 
Moreover, UEA and LRA in general have longer training sequence length, which would signalize Primal-Attention's efficiency. 
In contrast, the backbones on D4RL, WikiText-103 and ImageNet have more layers where canonical self-attention layers are the majority structures in Primal.$+$ as shown in Table~\ref{tab::arch} in this material. 
Besides, the efficiency gain is less significant also due to the shorter training sequence lengths on these datasets. 

\begin{table}[t]
\caption{Architecture of Primal.$+$ on different datasets.}
\label{tab::arch}
\begin{center}
	\resizebox{0.73\textwidth}{!}{
		\begin{tabular}{cccc}
			\toprule
			Primal.$+$  & {\tt canonical\_layer}+{\tt{[}primal\_layer{]}} & {\tt num\_head} & {\tt head\_dim} 
			\\ \midrule
			UEA & 1+{[}1{]} & 8 & 64        
			\\
			LRA & 1+{[}1{]} & 2 & 32        
			\\
			D4RL & 2+{[}1{]} & 4 & 64        
			\\
			WikiText-103 & 5+{[}1{]} & 8 & 64        
			\\
			ImageNet & 11+{[}1{]} & 6 & 64     
			\\ \bottomrule
	\end{tabular}}
\end{center}
\vspace{-5mm}
\end{table}

\section{Broader Impacts} \label{sec::broader_impact}

\paragraph{Societal Impacts}
In this work, we provide a new perspective to interpret self-attention through a KSVD problem with asymmetric kernels under the LSSVM framework. 
Compared to the canonical Transformers, our method is more efficient in tackling long-sequence datasets with our more efficient architectures that avoids the kernel matrix computation and also regularize the model with improved low-rank properties.
In this aspect, our method is more energy friendly as it can decrease the power consumption during training.

\paragraph{Possible Future Works}
We introduce a new self-attention mechanism from the primal perspective of the KSVD problem where feature maps are utilized rather than the kernels.
We currently work on the feature map corresponding to the Cosine similarity kernel in the paper that achieves state-of-the-art performances on the evaluated benchmarks. 
For more general setups and applications,  different feature maps and backbone architectures can be further investigated.
Therefore, it can extend our method to a wider range of tasks and possibly gain better performance under practical scenarios. 
These can be possible directions for future work.

\end{document}